\documentclass[conference]{IEEEtran}
\usepackage{times}

\usepackage[numbers]{natbib}
\usepackage{multicol}
\usepackage{amssymb}
\usepackage{graphicx}
\usepackage{xcolor}
\usepackage{booktabs}
\usepackage[bookmarks=true]{hyperref}
\let\labelindent\relax
\usepackage{enumitem}
\usepackage[symbol]{footmisc}
\usepackage{subcaption} 
\usepackage[font=small,labelfont=bf]{caption}

\newcommand{\model}[0]{HACMan++}
\newcommand*\rebuttal{\textcolor{black}} 
\newcommand{\rebuttalm}[1]{\leavevmode\color{black}{#1}}

\title{HACMan++: Spatially-Grounded Motion Primitives for Manipulation}

\author{Bowen Jiang*$^{1}$, Yilin Wu*$^{1}$, Wenxuan Zhou$^{1}$, Chris Paxton$^{2}$, David Held$^{1}$ \\
$^{1}$CMU, $^{2}$AI at Meta

}

\begin{document}

\setcounter{figure}{1}
\makeatletter
\let\@oldmaketitle\@maketitle
\renewcommand{\@maketitle}{\@oldmaketitle
    \vspace{0.5cm}
  \begin{center}
    \includegraphics[width=\linewidth]{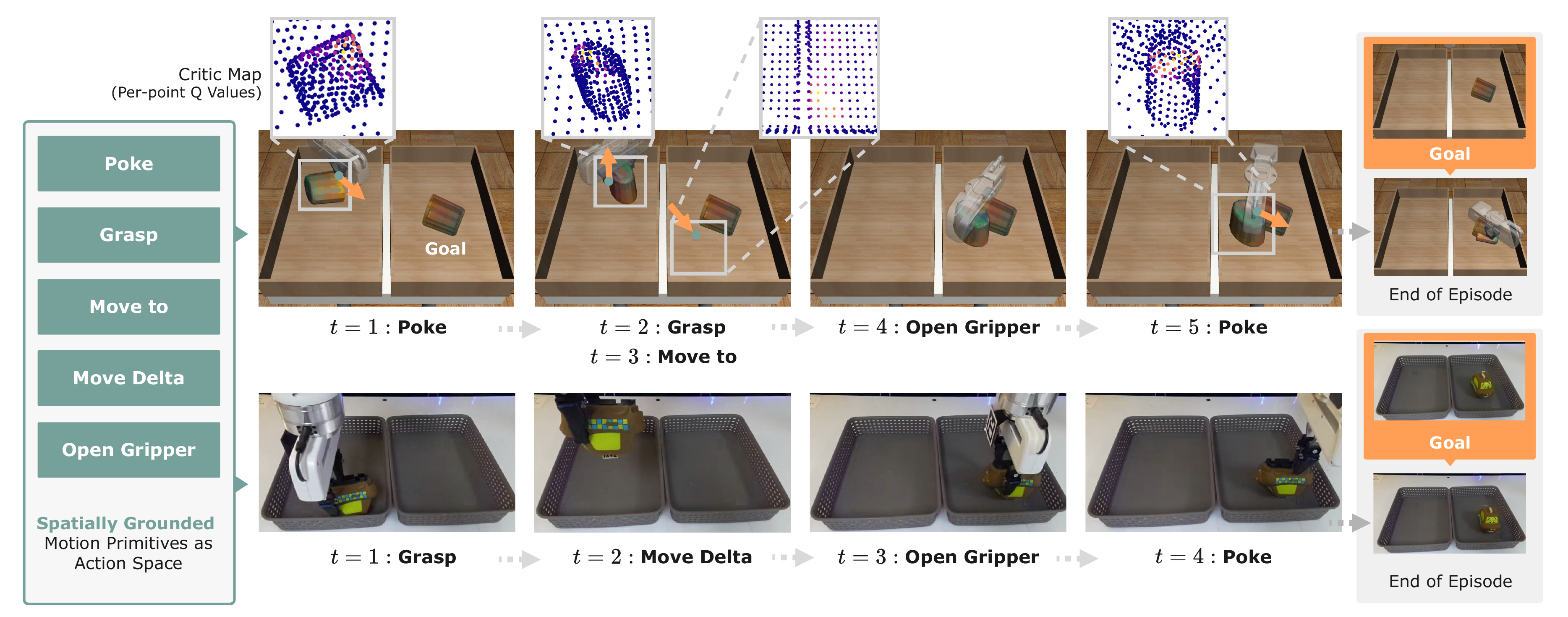}
  \end{center}
  \small{
  \textbf{Fig.~\thefigure:\label{fig:enter-label}}~
    Our method consists of a library of parameterized, \emph{spatially-grounded} motion primitives (left), consisting of a primitive type, primitive  location (where the primitive will be grounded), and primitive parameters.  These three components form the action space for a policy that we train with reinforcement learning.  Our method learns to select a sequence of primitives (and their corresponding locations and parameters) to perform a long-horizon manipulation task.  In the task shown here, the object is placed in one bin in an initial pose, and it must be moved into a second bin in a target pose. 
 At the top, we visualize the spatial grounding for the selected primitive; for each point we visualize the learned Q-value of selecting that point in the form of heatmaps as the grounding location for each primitive.}
  }
  \vspace{0.5cm}
  \medskip
\makeatother

\maketitle
\footnotetext[1]{detonates equal contribution}

\begin{abstract}
Although end-to-end robot learning has shown some success for robot manipulation, the learned policies are often not sufficiently robust to variations in object pose or geometry. To improve the policy generalization, we introduce spatially-grounded parameterized motion primitives in our method \model{}. Specifically, we propose an action representation consisting of three components: 
\textit{what} primitive type (such as grasp or push) to execute, \textit{where} the primitive will be grounded (e.g. where the gripper will make contact with the world), and \textit{how} the primitive motion is executed, such as parameters specifying the push direction or grasp orientation. 
These three components define a novel discrete-continuous action space for reinforcement learning. Our framework enables robot agents to learn 
to chain diverse motion primitives together and select appropriate primitive parameters to complete long-horizon manipulation tasks.
By grounding the primitives on a spatial location in the environment, our method is able to effectively generalize across object shape and pose variations. Our approach significantly outperforms existing methods, particularly in complex scenarios demanding both high-level sequential reasoning and object generalization. With zero-shot sim-to-real transfer, our policy succeeds in challenging real-world manipulation tasks, with generalization to unseen objects. Videos can be found on the project website:
\url{https://sgmp-rss2024.github.io}.
\end{abstract}

\IEEEpeerreviewmaketitle

\section{Introduction}

Despite recent progress in training manipulation policies with reinforcement learning (RL), it remains challenging to scale RL training to longer-horizon problems with broader task variations~\cite{gupta2019relay, lee2021ikea, zhou2023hacman, pateria2021hierarchical}.
A significant limitation is that most robot manipulation policies reason over the space of granular robot-centric actions, such as gripper or joint movements~\citep{mu2021maniskill, yu2019meta, robosuite2020, zhou2022ungraspable}.
These action spaces are highly inefficient for longer-horizon tasks due to exploration, credit assignment, and training stability challenges in deep reinforcement learning~\citep{gupta2019relay, pateria2021hierarchical}.

Instead of learning policies over low-level timesteps, 
the robot should reason about long-horizon manipulation problems with general, reusable primitives. 
For example, to make coffee, the robot may segment the task into picking up a mug and then placing it under the coffee machine.
This process involves decomposing the task into 
a ``grasping'' stage followed by a ``placing'' stage. With a similar idea, prior work has proposed applying a hierarchical structure in robot decisions, such as options or skill primitives~\citep{dalal2021accelerating, nasiriany2022augmenting,xiong2018parametrized}. These methods decouple the high-level decisions of ``what'' to do from the low-level decisions of ``how'' to execute robot motions.
However, our experiments demonstrate that this prior work on using skill primitives shows limited generalization across different object geometries and poses.

We desire a model that can both reason over temporal abstractions (i.e. reasoning about a sequence of parameterized skill primitives) as well as achieve object pose and shape generalization. 

In this work \model{}, 
we propose to learn manipulation policies with RL using a set of \emph{spatially grounded} motion primitives. The motion primitives consist only of basic manipulation motions such as grasping, placing, or pushing. Each primitive is parameterized by a location selected from the observed point cloud, which the primitive is defined relative to, and a vector of additional parameters defining the details of the gripper motion. For example, ``grasping'' is parameterized by a location on the object point cloud to grasp and a gripper orientation; ``placing'' selects a location on the background point cloud and a gripper orientation; ``pushing" selects a contact location and a push direction. We also include two ``move" primitives to allow for more generic robot motions.

To train a reinforcement learning policy with this action space, we leverage hybrid actor-critic maps~\cite{zhou2023hacman,feldman2022hybrid}. 
Given a 3D object point cloud, our method trains a critic to output per-point, per-primitive scores, which form a
primitive-conditioned ``Critic Map."  Our method selects the best primitive and the corresponding location with the highest score in the primitive-conditioned critic map.
Compared to previous work on 3D hybrid actor-critic maps~\cite{zhou2023hacman} which is limited to one non-prehensile poking skill, we include a comprehensive set of heterogenous primitives to enable the robot to perform a wider variety of tasks. Another related line of work~\cite{feldman2022hybrid} is demonstrated on only a single task and four objects, whereas we demonstrate our approach on six different tasks and a wide variety of object geometries.

\rebuttal{The contributions of our paper include:
\begin{enumerate}
    \item A set of \textbf{diverse and generic spatially-grounded motion primitives} that can solve a range of complex tasks that could not be solved by prior work. 
    \begin{itemize}
        \item Compared to prior work that uses diverse motion primitives ~\citep{dalal2021accelerating, xiong2018parametrized}, our primitives are spatially-grounded and outperform prior work.
        \item Compared to prior work that uses specially-designed spatially-grounded primitives for a single task~\citep{feldman2022hybrid,zhou2023hacman}, our primitive set is more generic and applies to a wide range of tasks.
    \end{itemize}  
    \item An RL training framework that incorporates the primitive selection and spatial-grounding selection using the critic.
\end{enumerate}}

\rebuttal{Our experimental contributions include:
\begin{enumerate}
    \item We demonstrate that our method learns complex skills that generalizes over unseen objects, achieving an 89.5\% success rate on training objects and an 84.9\% success rate on unseen object categories on our Double Bin task.
    \item We show that our method significantly outperforms prior work that includes diverse primitives that are not spatially-grounded on diverse simulation tasks.
    \item We also perform real robot experiments for a DoubleBin object pose alignment task, which achieves 73\% success rate.
\end{enumerate}}

In addition to our main experiments, we also show preliminary results of extending the concept of spatially grounded motion primitives to dexterous hand manipulation tasks in Appendix~\ref{appendix: simulation: additional tasks}, demonstrating the potential for this approach to generalize to other robot morphologies.

\section{Related Work}

\textbf{Hierarchical Reinforcement Learning.} Prior work has integrated a hierarchical structure into reinforcement learning 
to reduce the challenge of long-horizon reasoning for RL algorithms~\citep{sutton1999between}. In hierarchical reinforcement learning, a high-level policy will communicate with one or more low-level policies to finish the task. However, it can be difficult to jointly optimize both the high and low-level policies~\citep{haarnoja2018latent}. Alternatively, prior work has proposed to first learn a set of low-level skills from an \rebuttal{offline} dataset~\citep{pertsch2021accelerating,ajay2020opal,Shankar-2020-126755, Shankar-2020-126754, singh2020parrot}. 
Instead, we follow prior work in the robotics domain and define the low-level policies as commonly used primitives such as grasping, placing, and pushing~\citep{dalal2021accelerating, nasiriany2022augmenting}. We compare our method to other methods that use ``skill libraries," including some hierarchical RL methods, explained below.

\textbf{Skill Libraries.} \rebuttal{Prior work~\citep{joha2019} has specifically designed a set of primitives including approach, contact, fit, align, and insertion, as a skill library. However, this set of primitives is not generalizable to other tasks. Furthermore, it assumes a pre-specified order of primitives to be executed to finish a given task. Another line of work defines the action space of the RL policy based on a more general set of pre-defined parametrized primitives such as RAPS~\citep{dalal2021accelerating}, MAPLE~\citep{nasiriany2022augmenting}, and Parameterized DQN~\citep{xiong2018parametrized}. The RL policy learns to  automatically chain different primitives together to achieve a task without assuming a fixed order of primitives. This also means that the agent can re-execute primitives when a failure occurs. Our method inherits the benefits of those work in RL policy with parameterized primitives and also differs from them in that we spatially ground the primitives to improve spatial reasoning. We compare our method to these prior methods in the experiments and demonstrated significantly improved performance.}

\begin{figure*}[ht]
    \centering
    \includegraphics[width=\linewidth]{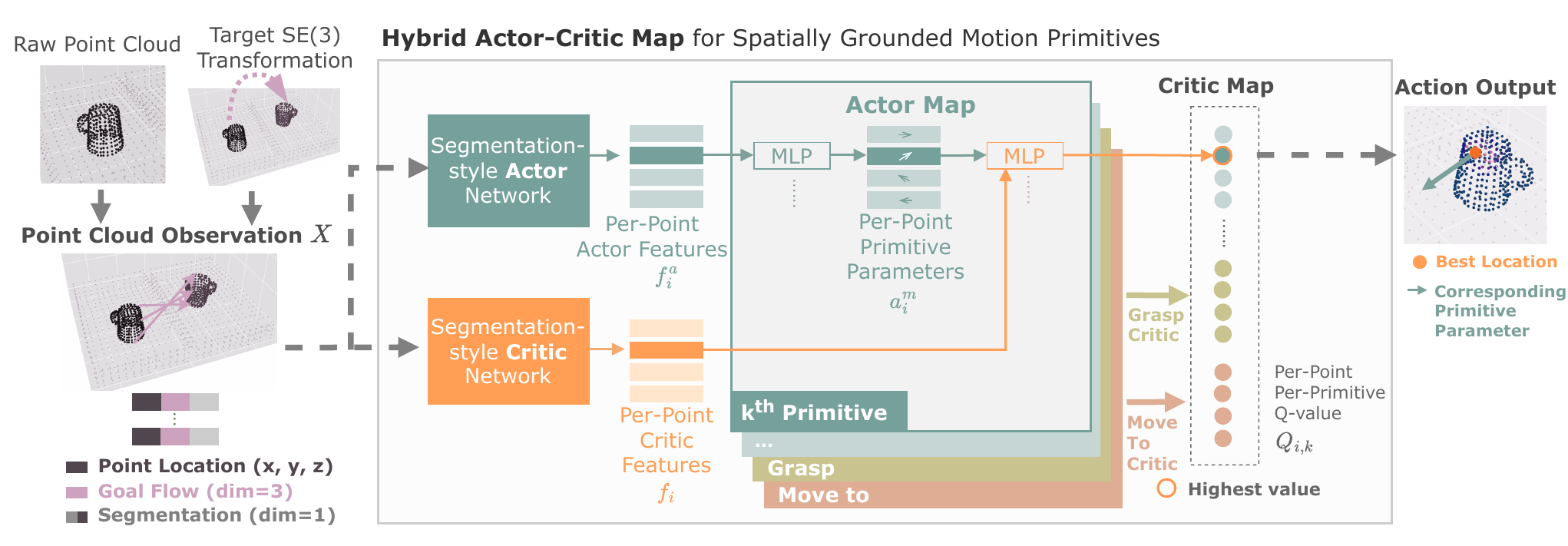}
    \caption{Our method processes a point cloud to estimate a set of per-point primitive parameters $a_i^m$ for each point $x_i$ in the point cloud and for each primitive in our primitive set. We then compute a set of ``Critic Maps" (one per primitive) which estimate the Q-value $Q_{i,k}$ of using each primitive $k$, grounded at each point $x_i$, and parameterized by the estimated primitive parameters $a_i^m$.  We either sample from the Critic Map (during training) or choose the point and primitive with the highest score (during evaluation) for robot execution.}
    \label{fig:architecture}
\end{figure*}
\textbf{Spatial Action Maps.} Spatial action maps connect a dense action representation with visual inputs using segmentation-style models~\citep{zeng2020transporter, Seita2022toolflownet, shridhar2022peract, Mo_2021_ICCV,xu2022umpnet,wu2020spatial}. Our method proposes a novel combination of motion primitives with spatial action maps to incorporate both temporal abstraction and spatial reasoning. Most prior work on spatial action maps has limitations on requiring expert demonstrations for imitation learning~\citep{zeng2020transporter, Seita2022toolflownet, shridhar2022peract} or is limited to one-step decisions without sequential reasoning~\citep{Mo_2021_ICCV,xu2022umpnet,wu2020spatial}. 
Our work is the most related to~\citep{zhou2023hacman, feldman2022hybrid}; 
however, \citet{zhou2023hacman} is limited to one non-prehensile skill (pushing) while we use a set of heterogeneous skills that can be combined to achieve more complex tasks. \citet{feldman2022hybrid} uses 2 skills, grasp and shift, and their horizon is limited to 2 (shift and then grasp), whereas our method allows the algorithm to chain the skills together in different sequences as appropriate for different tasks. Further, \citet{feldman2022hybrid} demonstrates their method on a single task with four object types, whereas we demonstrate our method on six different tasks and a wide variety of object shapes. \rebuttal{More detailed discussions between our method and~\citep{zhou2023hacman, feldman2022hybrid} are included in Appendix~\ref{appendix: extended discussion with related work}.}

\section{Background}
A stochastic sequential decision problem can be formalized as a Markov Decision Process (MDP), characterized by $(S, A, P, r, \gamma)$. Here, $S$ denotes the states, $A$ represents the actions, $P(s_{t+1} | s_t, a_t)$ is the likelihood of transitioning from state $s_t$ to state $s_{t+1}$ given action $a_t$, and $r(s_t, a_t, s_{t+1})$ is the reward obtained at time $t$. The goal within this framework is to optimize the return $R_t$, which is the sum of discounted future rewards, expressed as $R_t=\sum_{i=0}^{\infty}\gamma^{i}r_{t+i}$. Under a policy $\pi$, the expected return for taking action $a$ in state $s$ is described by the Q-function $Q^{\pi}(s,a)=\mathbb{E}_\pi[R_t|s_t = s, a_t = a]$.

\model{} leverages Q-learning-based algorithms for continuous action spaces~\citep{lillicrap2015continuous, fujimoto2018addressing}. These methods are characterized by a policy $\pi_\theta$ with parameters $\theta$, and a Q-function $Q_\phi$ with parameters $\phi$. Training involves collecting a dataset $D$ of state transitions $(s_t, a_t, s_{t+1})$, with the Q-function's loss formulated as:
\begin{equation}
L(\phi) = \mathbb{E}_{(s_t,a_t, s_{t+1})\sim D}[(Q_\phi(s_t,a_t) - y_t)^2],
\label{eqn:bellman}
\end{equation}
with $y_t$ being the target value, determined by:
\begin{equation}
y_t = r_t + \gamma Q_{\phi}(s_{t+1}, \pi_{\theta}(s_{t+1})).
\end{equation}
The optimization of the policy $\pi_\theta$ is described by the loss function:
\begin{equation}
J(\theta) = -\mathbb{E}_{s_t\sim D}[Q_{\phi}(s_t, \pi_\theta (s_t))].
\end{equation}

\section{Method}
\rebuttal{\textbf{Assumptions.} We assume that the robot agent records a point cloud observation of a scene $\mathcal{X}$, which may be obtained from one or more calibrated depth cameras. We further assume that this point cloud is segmented into object $\mathcal{X}^{obj}$ and background $\mathcal{X}^{b}$ components. See Appendix~\ref{appendix:algorithm} for details.}

To address the challenges of long-horizon manipulation tasks, our method uses a set of parameterized motion primitives, and learns how to both 1) chain these primitives together to achieve a task and 2) select appropriate parameters for the execution of each primitive. Section~\ref{sec:Action Representation} defines the structure of the proposed action representation. Section~\ref{sec:Parameterized Motion Primitives} lists the specific choices of parameterized motion primitives. Section~\ref{sec:Hybrid RL Algorithm} describes how we train the policy with the proposed action space with the RL algorithm.

\subsection{Action Representation}
\label{sec:Action Representation}

Our action representation comprises three key elements: the primitive type $a^{prim}$, the primitive location $a^{loc}$, and the primitive parameters $a^{m}$. These components collectively define the ``What'', ``Where'', and ``How'' of each sequential skill execution.

\vspace {1ex}

\noindent \textbf{Primitive Type \boldmath{$a^{prim}$}} determines the type of primitive the robot will execute, such as poking, grasping, or placing (see the full list in Section~\ref{sec:Parameterized Motion Primitives}). The robot policy aims to learn to perform different tasks by chaining the primitives in appropriate order based on the observations. Each type of primitive is uniquely parameterized to allow for variations in execution, adapting to the specific demands of the task. Once the parameters are specified, these primitives are executed with a low-level controller. 

\vspace {1ex}

\noindent \textbf{Primitive Location \boldmath{$a^{loc}$}} is a selected point of interaction in the scene, chosen from the observed point cloud $\mathcal{X}$. The selected point \emph{grounds} each primitive in the observed world: the robot action will be applied at a location \emph{relative to} the selected point, as defined by the primitive parameters $a^{m}$.

\vspace {1ex}

\noindent \textbf{Primitive Parameters \boldmath{$a^{m}$}} detail how the robot will execute the chosen primitive at the selected location $a^{loc}$. It includes aspects like gripper orientation while approaching the object, an offset with respect to the chosen primitive location, and post-contact movement. Details are primitive-type-dependent and are described below.

\subsection{Parameterized Motion Primitives}
\label{sec:Parameterized Motion Primitives}

We use five distinct and generic motion primitives, that collectively satisfy the needs of a wide range of manipulation tasks, following the primitive designs from previous work~\citep{dalal2021accelerating, nasiriany2022augmenting}. Each primitive has its own specific parameters described below. More details of the motion parameters for each primitive can be found in Appendix~\ref{appendix:algorithm}. 

\vspace {1ex}

\noindent \textbf{Poke:} This primitive applies a non-prehensile poking motion to the target object~\citep{zhou2023hacman, zeng2018learning, feldman2022hybrid, agrawal2016learning}. The robot moves the fingertip of the gripper to the selected primitive location $a^{loc}$ on the object as the initial contact point (see Appendix~\ref{appendix:algorithm} for details). The motion parameters $a^{m}$ consists of two parts: 1) the 2D gripper orientation while approaching the initial contact point, and 2) parameters that describe the poking motion after the gripper reaches the initial contact point on the object, defined as a 3D vector of gripper translation.

\vspace {1ex}

\noindent \textbf{Grasp:} This primitive grasps the target object and then lifts it up~\citep{mousavian20196, sundermeyer2021contact, feldman2022hybrid}. The primitive location $a^{loc}$ under the grasp primitive type defines a grasping point on the object. The motion parameters \rebuttal{$a^{m}$} detail the 2D gripper orientation while approaching the grasping point. Upon reaching the grasping point, the gripper closes to grasp the object. It then lifts up by a pre-specified distance (see Appendix~\ref{appendix:algorithm}). 

\vspace {1ex}

\noindent \textbf{Move to:} 

This primitive moves the gripper to a location that is defined relative to a point $a^{loc}$ selected from the background point cloud $\mathcal{X}^{b}$. The primitive parameters $a^m$ contain two parts: 1) the 2D gripper orientation when approaching the location, and 2) a 3D vector defining an offset from the selected location $a^{loc}$; the target point for the gripper to move to is given by the selected location $a^{loc}$ plus this offset. The selected location $a^{loc}$ grounds this motion 
on the point cloud, whereas the added offset gives the robot more flexibility in where to move. 
To speed up exploration, we restrict the primitive location $a^{loc}$ to be selected from the background points and we only execute this primitive when the gripper is already grasping an object.

\vspace {1ex}

\noindent \textbf{Open Gripper:} This primitive opens the gripper. The selected location $a^{loc}$ has no influence on the action, and this primitive does not require any motion parameters.

\vspace {1ex}

\noindent \textbf{Move delta:} To account for any nuanced movements that are difficult to achieve with the above primitives, we include the ``Move delta" primitive to move the gripper by a 3D delta movement and 2D orientation. Motion parameters for this primitive specify a delta translation and rotation of the gripper. We restrict this primitive to only be selected when the robot is already grasping an object. 

\begin{figure*}[ht]
    \centering
    \includegraphics[width=\textwidth]{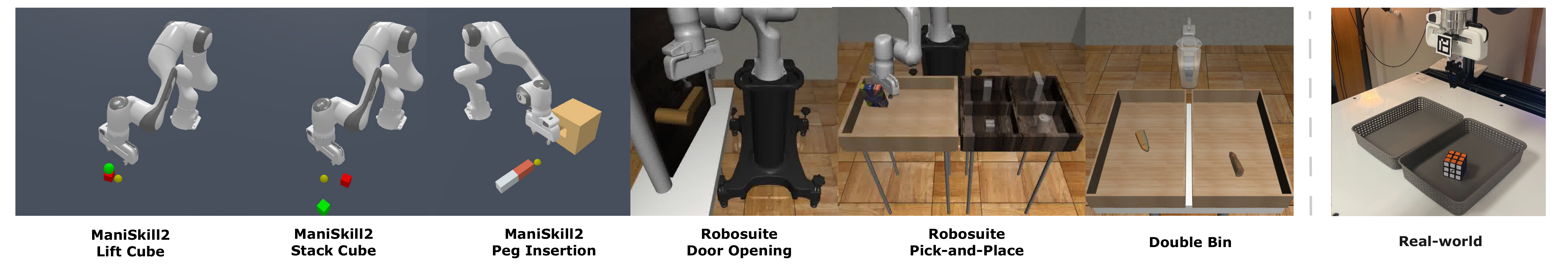}
    \caption{We evaluate our method on multiple object manipulation tasks that require picking, placing, and poking objects.  
    From left to right, we show the six simulation tasks: ManiSkill2 Lift Cube, ManiSkill2 Stack Cube, ManiSkill2 Peg Insertion, Robosuite Pick-and-Place, Robosuite Door Opening, and a customized Robosuite DoubleBin environment. We also show our real-world experiment setup which mimics the DoubleBin simulation environment. 
    }
    \label{fig:tasks}
\end{figure*}
\begin{figure*}[ht]
    \centering
    \includegraphics[width=\linewidth]{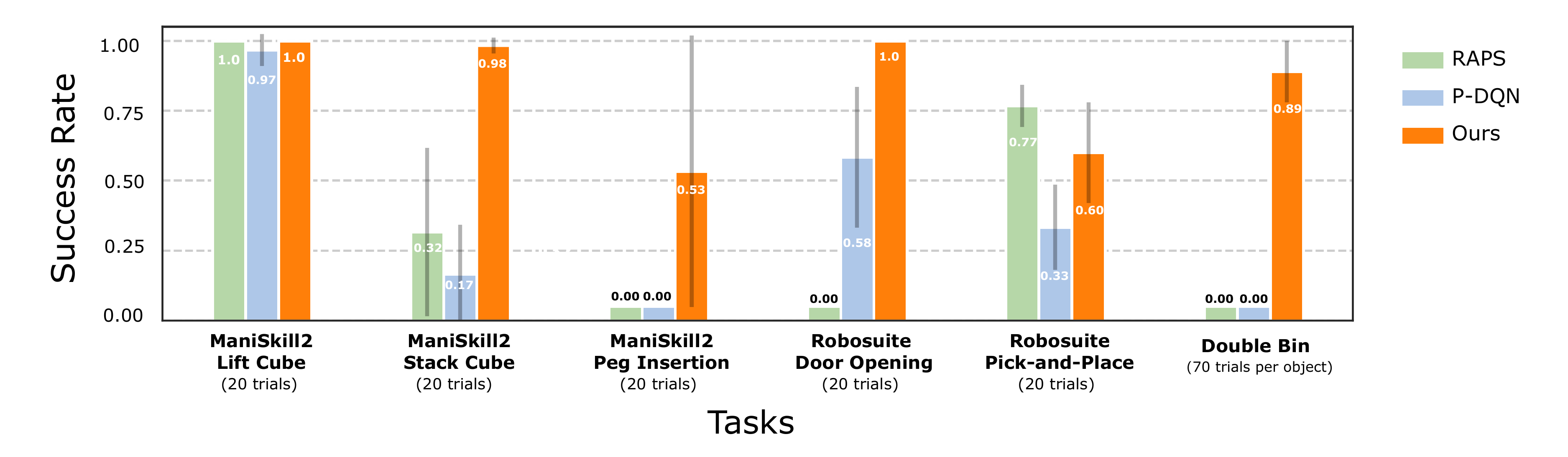}
    \caption{Performance of our method compared to baselines RAPS~\cite{dalal2021accelerating} and P-DQN~\cite{xiong2018parametrized} on six different tasks. \rebuttal{For all the \emph{ManiSkill} tasks and \emph{Robosuite} tasks, we report the success rate averaged over 20 trials. For \emph{DoubleBin} tasks, we report the average success rate over 32 objects, each tested with 70 trials.} These baselines use the same skill primitives as our approach but they are not spatially grounded, e.g. they do not ground the primitives on a point selected by the policy from the observed point cloud.  }
    \label{fig:evals_baselines}
\end{figure*}

\subsection{Hybrid RL Algorithm}
\label{sec:Hybrid RL Algorithm}
\model{} integrates a multi-primitive approach with existing Q-learning-based RL algorithms~\citep{fujimoto2018addressing, haarnoja2018soft, lillicrap2015continuous}.
Our action space includes 
both discrete and continuous components:
the primitive type $a^{prim}$ is selected from K primitives; for each primitive type, the primitive location $a^{loc}$ is selected from N points from the observed point cloud; whereas the motion parameters $a^{m}$ are a vector of continuous values.

The overall architecture of our approach is shown in Figure~\ref{fig:architecture}. The agent receives as input a point cloud observation of size N. We first use a segmentation-style point cloud network to output per-point actor features $f_i^a$ for each point $x_i$. These features are shared across the $K$ different primitives. We then input each of these features into a per-primitive MLP to output motion parameters $a^{m}_{i, k}$ for each point $x_i$ and each primitive $k$. We refer to these outputs as an ``Actor Map".

Our method also extracts per-point critic features $f_{i}$ for each point $x_{i}$  through a segmentation-style point cloud critic feature extractor. These features are shared across the $K$ different primitives. The per-point motion parameters $a^{m}_{i, k}$ are then concatenated with per-point critic features $f_i$ and input into \rebuttal{a multi-layer perceptron (MLP)} to calculate per-point Q values $Q_{i, k}$ for each point $x_i$ and each primitive $k$; this Q-value represents the effectiveness of executing the $k^{th}$ primitive with the motion parameters $a^{m}_{i, k}$ at the primitive location $x_i$. The above procedure generates a  ``Critic Map" with a total of $K × N$ Q-values across all points and all primitives (see Figure~\ref{fig:architecture}).

The optimal action is chosen by selecting the highest Q-value $Q_{i, k}^{max}$ from the critic map, which corresponds to
primitive type $k$, 
primitive location $x_i$, 
and motion parameters $a^{m}_{i, k}$. 
During training, the policy selects primitive types and locations by sampling from a softmax distribution over Q-values to balance exploration and exploitation, formalized as:

\begin{equation}
\pi^{discrete}(k, x_{i} \mid s)= \frac{\exp(Q_{i,k} \slash \beta)} {\sum_{k=1} ^ K \sum_{i=1} ^ N \exp(Q_{i,k} \slash \beta)}
\label{eqn:pi_loc}
\end{equation}
where $\beta$ is a temperature parameter modulating the softmax function, guiding the agent's exploratory behavior. 

The Q-function is updated according to the Bellman equation (Equation~\ref{eqn:bellman}) following TD3~\cite{fujimoto2018addressing}.
To update the primitive parameters $a_{i, k}^m$, we similarly follow the TD3 algorithm~\cite{fujimoto2018addressing}: If we define the actor $\pi^{\theta}_{i,k}(s)$ as the function parameterized by $\theta$ that maps from the observation $s$ to the action parameters $a_{i, k}^m$ for a given point $x_i$ and primitive $k$, then the loss function for this actor is given by:
\begin{equation}
J(\theta) = -Q_\phi(f_{i}, a_{i, k}^m) = -Q_\phi(f_{i},\pi^{\theta}_{i,k}(s)),
\end{equation} 
where $Q_\phi$ is the critic network and $f_{i}$ is the critic feature of the point $x_i$.  

To assist the network in understanding the relationship between the observation and the goal, we compute the correspondence between the points in the observation and the points in the goal (see Appendix~\ref{appendix:algorithm} for details).  For every point in the observation, we append to the input a 3-dimensional vector indicating the delta to its corresponding goal location, which we refer to as ``goal flow" (see Figure~\ref{fig:architecture}).

\section{Experimental Setup}

We evaluate our method on three ManiSkill tasks (Sec.~\ref{sec:ManiSkill Tasks}), two Robosuite tasks (Sec.~\ref{sec:Robosuite Task}), as well as a DoubleBin task (Sec.~\ref{sec:DoubleBin Tasks})
as illustrated in Figure~\ref{fig:tasks}. This section outlines the setup, objective, and reward function for each task.

\subsection{ManiSkill Tasks}
\label{sec:ManiSkill Tasks}

We evaluate our method with three tasks from ManiSkill~\citep{mu2021maniskill} (Figure~\ref{fig:tasks}, Left). For these tasks, we train the hybrid actor-critic map with the default reward functions defined in the ManiSkill benchmark
~\citep{mu2021maniskill}.

\vspace {1ex}

\noindent \textbf{Lift Cube:} The agent is tasked with picking up a cube and lifting it to a specified height threshold. The initial cube position and orientation are randomized.

\vspace {1ex}

\noindent \textbf{Stack Cube:} This task involves stacking a red cube on top of a green cube, requiring precision in alignment. The initial position and orientation of both cubes are randomized.

\vspace {1ex}

\noindent \textbf{Peg Insertion:} This task involves inserting a peg horizontally into a hole in a box. 
As the original ManiSkill paper~\citep{mu2021maniskill} reports a 0 success rate on this task, we slightly simplify this task by removing the variations in both the hole's location and the peg's initial pose as well as marginally increasing the clearance of the hole. We compare to baseline approaches with these same environment modifications.

\subsection{Robosuite Task}
\label{sec:Robosuite Task}
We also evaluate our method with two tasks from Robosuite~\citep{robosuite2020}(Figure~\ref{fig:tasks}, Left). For these tasks, we train with the default dense reward functions in the Robosuite benchmark~\citep{robosuite2020}.

\vspace {1ex}
\noindent \textbf{Pick-and-Place}: The task is initialized with one object at a random position in a large single bin and the goal is to place the object into a specified small container on the side. There are four containers in total and four objects, including cube, box, can and milk carton.  
\vspace {1ex}

\noindent \textbf{Door Opening}: A door with a handle is placed in front of a single robot arm in this task. The agent needs to learn to rotate the door handle and open the door. 

\begin{figure*}[th]
    \centering
    \includegraphics[width=\linewidth]{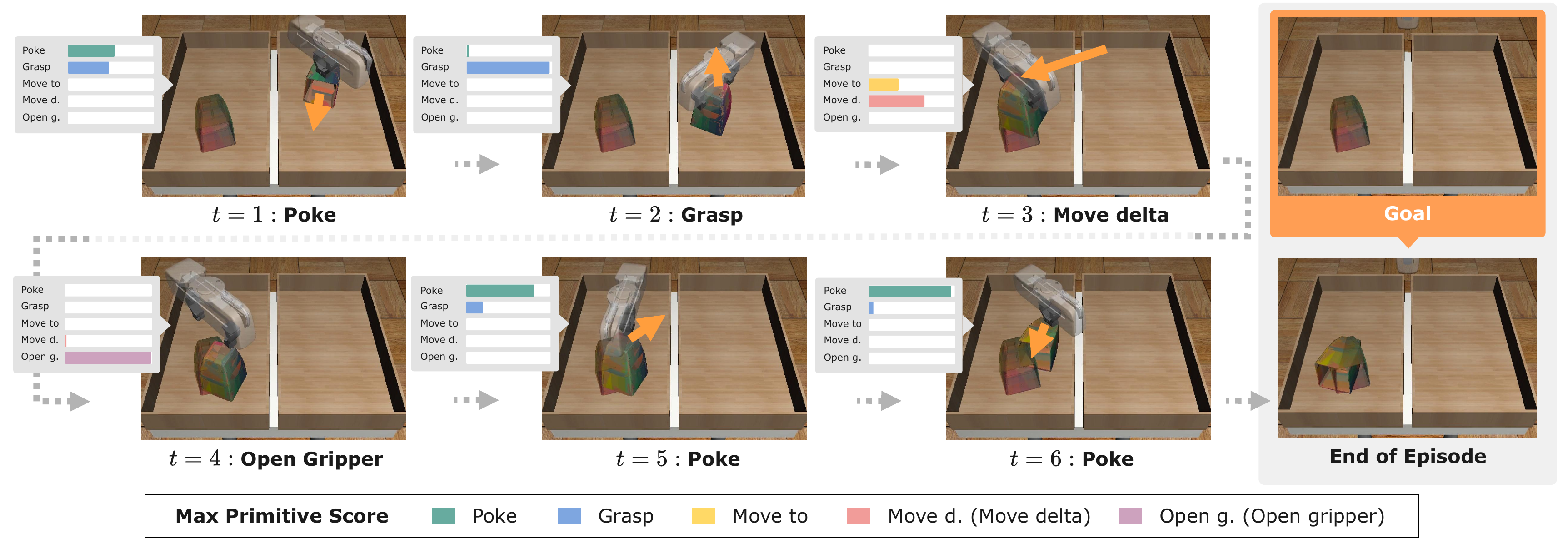}
    \caption{A simulation rollout of our policy.  The goal is shown in the top right, and also overlayed on each observation.  At each step, we visualize the scores that we assign to each of the primitives.  We also visualize the selected primitive location and parameters (orange arrow). As shown, our method learns to chain a sequence of grounded primitives to accomplish a challenging long-horizon manipulation task.}
    \label{fig:rollout}
\end{figure*}

\subsection{DoubleBin Task}
\label{sec:DoubleBin Tasks}

To further demonstrate the benefits of spatial grounding, we design the DoubleBin task (Figure~\ref{fig:tasks}, Right). It is built on top of Robosuite~\citep{robosuite2020} with the Mujoco simulator~\citep{todorov2012mujoco}. Compared to the ManiSkill tasks, it requires longer horizon reasoning and has more object shape variations. Each episode starts with two bins with one object in a randomly selected bin. The objective of the robot agent is to perform a sequence of motions to manipulate the object to a pre-specified 6D goal pose in the opposite bin. This resembles a common scenario in warehouse automation and assembly lines. The reward function is the average norm of the point cloud correspondence vectors between the object's current state and goal state state (see Appendix~\ref{appendix:simulation environments}).


At each episode, we sample one object from a set of 32 objects with diverse geometries for training. The agent needs to dynamically adapt its manipulation strategies to suit the unique geometry of each object. 
We evaluate our method on 7 unseen object instances (from training object categories) and 5 objects from unseen object categories.

\section{Simulation Results}
In our simulation experiments, we aim to answer the following questions:
\begin{itemize}
    \item Do spatially grounded primitives enable better performance in high precision tasks than previous methods?
    \item Does our method reasonably select appropriate primitives at each step from a set of primitives and strategically compose them together to solve long-horizon tasks?
    \item Does the learned policy generalize to unseen objects?
\end{itemize}
The comparison between our method and baselines over six tasks is reported in Figure~\ref{fig:evals_baselines}. The details of the training and evaluation procedures can be found in Appendix~\ref{appendix:simulation environments}.

\vspace {1ex}
\begin{table}[]
\centering
\caption{Differences Between Our Method and  Baselines.}
\begin{tabular}{@{}lcc@{}}
\toprule
 & \textbf{Spatially Grounded} & \textbf{Primitive Selection} \\ \midrule
Ours & $\checkmark$ & Argmax of Critic Scores \\
P-DQN~\citep{xiong2018parametrized} & $\times$ & Argmax of Critic Scores \\
RAPS~\citep{dalal2021accelerating} & $\times$ & Argmax of Actor Probabilities \\ \bottomrule
\end{tabular}
\label{tab:action_representation_feature_comparisons}
\end{table}

\begin{table*}
\caption{\textbf{Generalization to Unseen Objects.} \model{} shows strong generalization to previously-unseen instances of classes in the training data, and even generalizes well to unseen object categories.}
\centering
\begin{tabular}{@{}lcccc@{}}
\toprule
\textbf{Object Set Split} & \textbf{Success Rate (10 steps)} & \textbf{Success Rate (20 steps)} & \textbf{Success Rate (30 steps)} & \textbf{\# of Objects} \\ \midrule
Train & 0.676 ± .010 & 0.845 ± .010 &0.892 ± .010 &  32 \\
Train (Common Categories) & 0.746 ± .020 & 0.903 ± .016 & 0.937 ± .011&  13 \\
Unseen Instance (Common Categories)&  0.737 ± .020 & 0.903 ± .023 & 0.952 ± .023 & 7 \\
Unseen Category & 0.601 ± .003& 0.784 ± .027 &0.849 ± .003 &  5 \\ \bottomrule
\end{tabular}
\label{tab:generalization_to_unseen}
\end{table*}

\noindent \textbf{Effect of Spatial Grounding.} 
To demonstrate the benefits of spatial grounding, we compare our method to two baselines, P-DQN and RAPS~\citep{dalal2021accelerating,xiong2018parametrized}. Both of the baselines use parameterized primitives as the action space of their RL policies, but the primitives are \textbf{not} spatially-grounded. For primitives that involve location parameters, both of the baselines directly regress the location parameters, instead of selecting a location from the observed point cloud as in our method. P-DQN selects primitives based on the critic scores of each primitive type (rather than the critic scores of each primitive type and \textit{location} in our method), while RAPS directly outputs both action probabilities and the primitive parameters from the actor. Table~\ref{tab:action_representation_feature_comparisons} highlights the differences between our method and these baselines. A more detailed description of the implementation of the baselines can be found in Appendix~\ref{appendix:algorithm}. 

The results are shown in Fig.~\ref{fig:evals_baselines}. Although these baselines perform well on the easiest task (\textbf{Lift Cube}), they struggle with the other tasks which require more precise spatial reasoning (\textbf{Stack Cube} and \textbf{Peg Insertion}) and/or generalization to object shape variations (\textbf{DoubleBin}). For Robosuite tasks, \emph{Pick-and-Place} 1) does not require precise placing since the goal can be at any position inside the container 2) and does not require generalization to object shapes because there are limited geometries (4 objects compared to 32 objects in the \emph{Double Bin} task). The \emph{Door Opening} task, on the contrary, requires more geometric reasoning so our method outperforms the baselines by a large threshold. In general, ours is the \emph{only} method that maintains reasonable performance across the six different tasks. Note that the manipulation tasks in our experiments require higher precision than the tasks reported in RAPS~\citep{dalal2021accelerating}. These results demonstrate the benefits of spatial grounding for precise manipulation tasks.

\begin{figure}[h!]
    \centering
    \includegraphics[width=\linewidth]{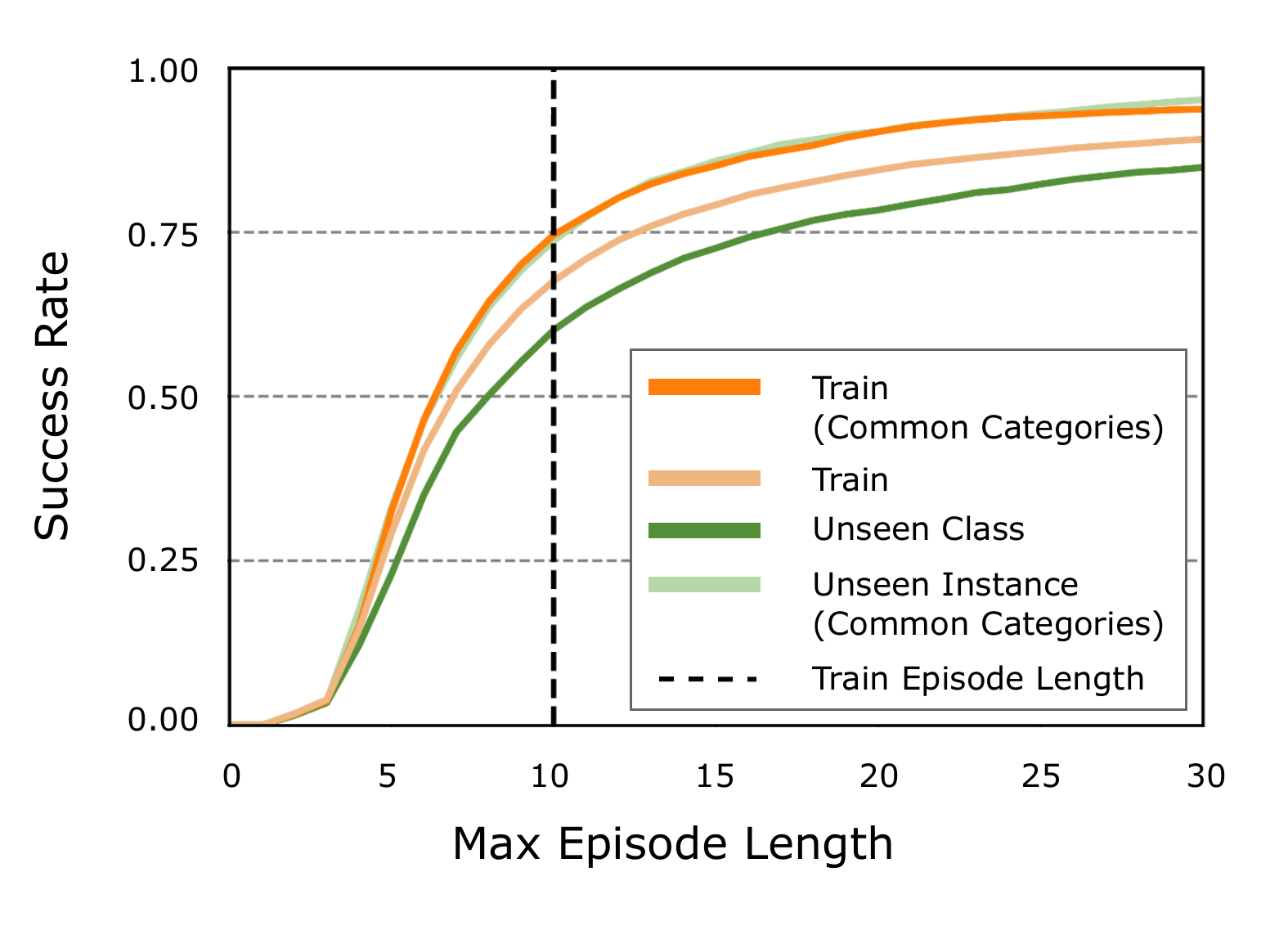}
    \caption{Success rate as a function of the episode length, for training objects (all), training objects from categories with many instances (common categories), unseen instances from those same common categories, and for unseen object classes. We train with an episode length of 10 but evaluate with varying episode lengths up to 30.}
    \label{fig:success_rate_vs_length}
\end{figure}

\vspace {1ex}

\noindent\textbf{Effect of Primitive Chaining}
The proposed method is able to chain primitives together in appropriate orders to solve different tasks, without requiring a pre-specified sequence of primitive types~\citep{simeonov2021long}. For example, in the DoubleBin task, our method learns to chain both the prehensile and non-prehensile primitives together with different orders under different situations. If the initial pose of the object is impossible for top-down grasping, our policy will first poke the object to a graspable pose as shown in the first step in Figure~\ref{fig:rollout}. After grasping, it also learns to use a move primitive (e.g. \emph{Move to} or \emph{Move delta}) to relocate the object to the other bin (e.g. step 3 in Figure~\ref{fig:rollout} and select \emph{Open Gripper} to release it. In some cases, the policy may perform a few \emph{Poke} primitives again to move the object into the correct pose if necessary (e.g., step 5-6 in Figure~\ref{fig:rollout}). This process of moving the objects across bins and into the correct poses is not possible without chaining the primitives strategically. Similarly, our method also demonstrates such strategic reasoning in ManiSkill Tasks, e.g., chaining grasp with multiple move primitives to complete the Peg Insertion Task.

\vspace {1ex}

\noindent \textbf{Generalization to Unseen Objects.} To demonstrate the generalization capabilities of the proposed method, we evaluate our model on the DoubleBin task with unseen objects, the results of which are summarized in Table~\ref{tab:generalization_to_unseen} and Figure~\ref{fig:success_rate_vs_length}. We report the performance averaged over $70 \times n$ trials, 
where n refers to the number of objects in the evaluation set. The overall success rate for achieving the target 6D goal transformation on the training objects is 89.2\% when the policy is evaluated with an episode length of 30. 

We evaluate the generalization capabilities of the model in three settings. First, we evaluate our method on unseen object instances. These evaluation objects are within the training object categories, but the exact object models are unseen. The unseen instances are randomly selected from the most common categories of objects from the full object dataset (for which there are many object instances), including plant container, salt shaker, pencil case, pill bottle, bottle, canister, and can. The performance of the model on these common object categories is 93.7\% for seen object instances. An evaluation on unseen instances from these same categories has nearly the same performance (95.2\%), demonstrating our model's ability to adapt to different object geometries within these categories. 

We also evaluate our method on objects in unseen categories that were not included in training (e.g., lunch bag) and achieve a success rate of 84.9\%, demonstrating the ability of our model to generalize to novel shapes.  A visualization of the training and unseen testing objects can be found in Appendix~\ref{appendix:simulation environments}.  

\vspace {1ex}

Furthermore, we conducted additional experiments with varying maximum episode length for evaluation; the results in Figure~\ref{fig:success_rate_vs_length} show that the success rate increases with a longer episode length. An episode length of 10 is used for training, so this figure also demonstrates the ability of our model to continue to improve performance beyond the training episode length. By changing from 10 steps to 30 steps, the performance is enhanced by at least 20\% for any object set. For objects in unseen category, the success rate can increase to 85\% when the maximum episode length increases to 30.

\vspace {1ex}

\noindent \textbf{Additional Experiments \& Analysis.} We also include some preliminary results of potential extension of our method to dexterous hand tasks in Appendix~\ref{appendix: simulation: additional tasks}. See  Appendix~\ref{appendix:simulation environments} for additional experiments and analysis.

\begin{figure*}[ht]
    \centering
    \includegraphics[width=\linewidth]{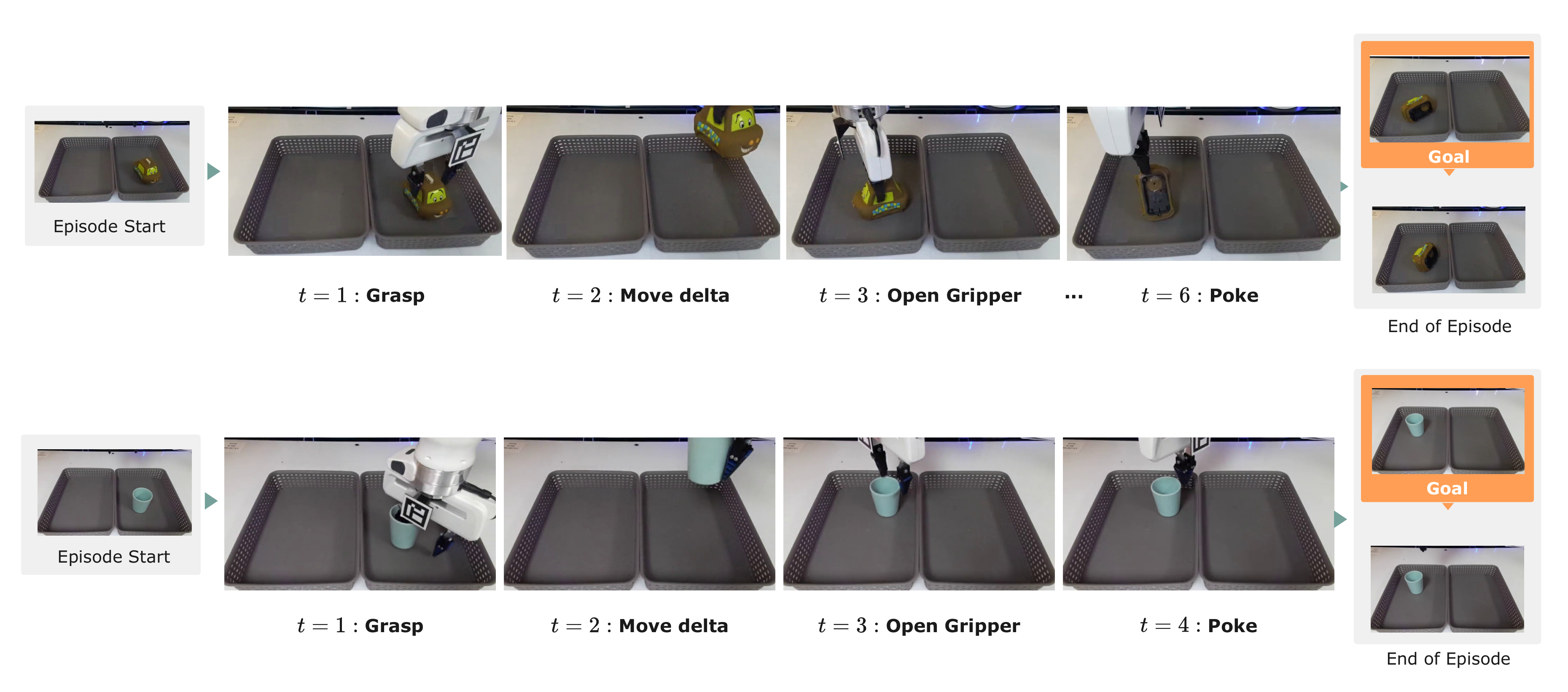}
     \vspace{-0.5cm}
    \caption{Two examples of real-world rollout of our policy. Our method learns to chain a sequence of actions to lift the object, move it across to the other bin, release the object, and then poke it to match the target pose more precisely. The first row shows the rollout of the car (toy) with a SE(3) goal. The second row shows rollout of the cup with a translation goal.  }
    \label{fig:real-rollout}
\end{figure*}

\section{Real-World Experiments}

We perform evaluations on the real world DoubleBin task with the policy trained in simulation as discussed in the previous sections. At the beginning of each episode, we place the object at a random pose in a randomly chosen bin. We also specify a goal SE(3) transformation, 
which can be either in the same bin as the initial object pose or in the opposite side bin. Among all the objects we are testing, Rubik's Cube, Bowl, Cup, Tennis are evaluated with the translation goals because of their rotation-symmetric shape. At each step, we perform point cloud registration to compute the correspondence from the current observation to the goal (see details in Appendix~\ref{appendix:real-world experiments}).

Similar to our simulation environment, an episode is deemed a success when the mean distance between each observation point on the object and its corresponding goal point is less than 3 cm.  We set a maximum episode length of 15 time steps (each time step corresponds to one primitive action).

\begin{figure}[]
    \centering
    \includegraphics[width=\linewidth]{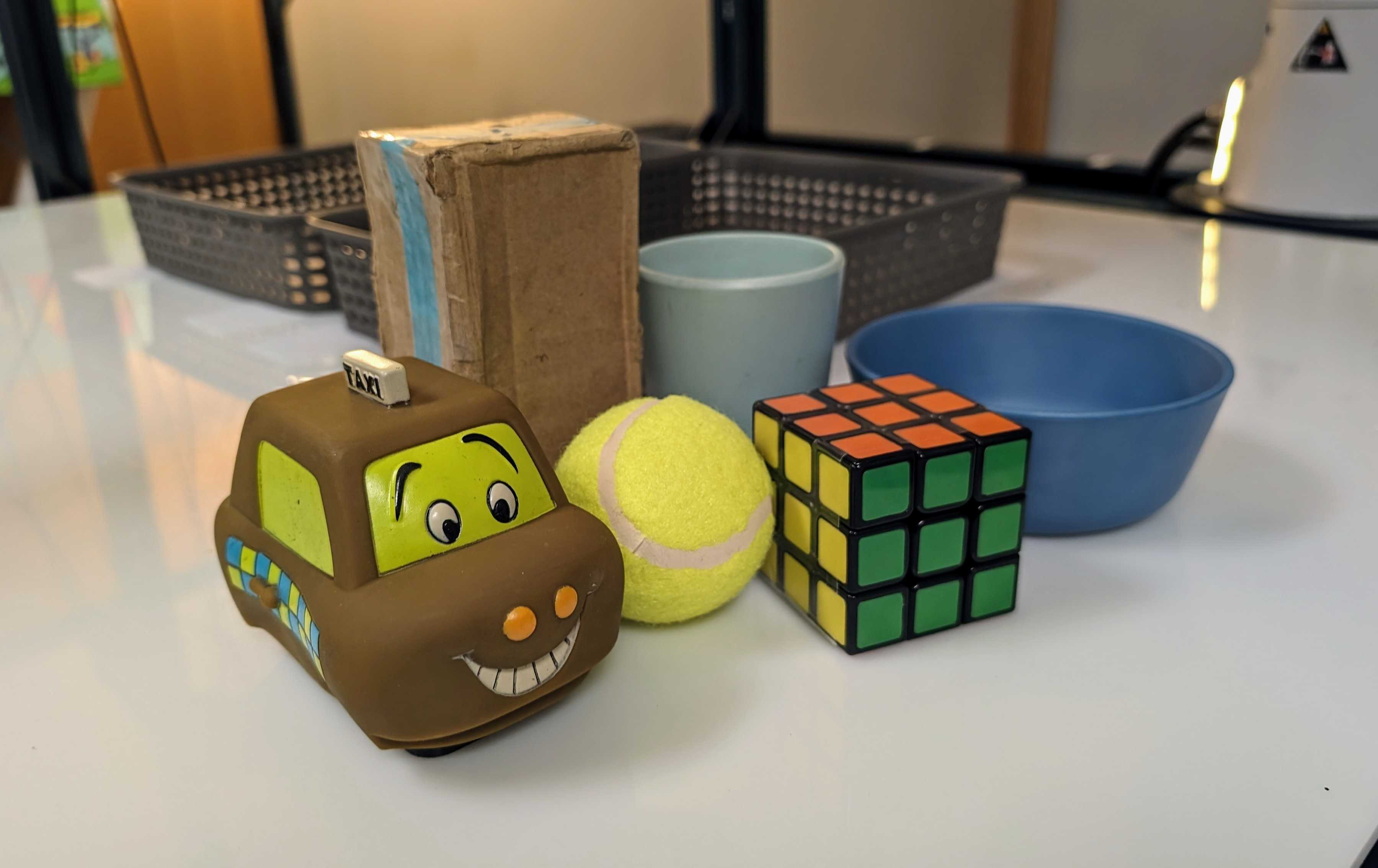}
    \caption{\textbf{Real-world Objects.} From left to right, the six objects are: \emph{Car (Toy)}, \emph{Cardboard}, \emph{Tennis}, \emph{Cup}, \emph{Rubik's Cube} and \emph{Bowl}.}
    \label{fig:real-world-objects}
\end{figure}

In our experiments, we select six objects with different materials and geometries, as shown in Figure~\ref{fig:real-world-objects}. Figure~\ref{fig:tasks} shows the real-world experiment setup. Figure~\ref{fig:real-rollout} demonstrates an example real-world trajectory rollout. Table~\ref{tab:real-world} shows the quantitative evaluation results. Our method is able to achieve an overall 73\% success rate.

\begin{table}[]
\caption{\textbf{Real-World Experiment Results}}
\centering
\begin{tabular}{@{}lccc@{}}
\toprule
\textbf{Object}     & \textbf{Same Side Goal} & \textbf{Opposite Side Goal} & \textbf{Subtotal} \\ \midrule
Rubik’s Cube        & 14/20                   & 12/20                       & 26/40             \\
Bowl                & 19/20                   & 16/20                       & 35/40             \\
Cup                 & 11/20                   & 14/20                       & 25/40             \\
Tennis              & 19/20                   & 19/20                       & 38/40             \\
Cardboard  & 11/20 &  15/20  & 26/40\\
Car (Toy)  & 12/20 & 14/20 &  26/40\\

\midrule
\textbf{Subtotal}   & 86/120                   & 90/120                       & 176/240           \\
\textbf{Percentage} & 72\%                    & 75\%                        & 73\%              \\ \bottomrule
\end{tabular}
\label{tab:real-world}
\end{table}

\section{Conclusion} 
\label{sec:conclusion}

In this work, we present spatially grounded motion primitives for robot manipulation tasks, leveraging hybrid actor-critic maps with reinforcement learning. Our agent learns to chain different spatially-grounded primitives with appropriately selected primitive parameters to complete a task. Our method adapts to diverse manipulation tasks and generalizes to diverse objects, succeeding in tasks that require both high-level sequential reasoning and low-level motion precision - where previous methods have fallen short. The effectiveness of our approach suggests the importance of primitives that are spatially grounded on points in the environment.  

\rebuttal{\textbf{Limitations.} Our approach to breaking down a manipulation task into motion primitives has shown adaptability across a range of scenarios; nonetheless, there are complexities in designing general primitives to accommodate every task. Although we have added some preliminary experiments exploring other gripper morphology (i.e. the dexterous Shadow hand task in Appendix~\ref{appendix: simulation: additional tasks}), more exploration is needed to determine the best way to apply spatially-grounded primitives to different gripper designs.
}
\section*{Acknowledgments}
We would like to thank Chialiang Kuo for his help to run experiments on Adroit environment. We also thank Zhanyi Sun and Ben Eisner for the insightful feedback throughout this project. This work is supported by National Institute of Standards and Technology under Grant No. 70NANB23H178. Any opinions, findings, and conclusions or recommendations expressed in this material are those of the author(s) and do not necessarily reflect the views of the NIST.
\bibliographystyle{plainnat}
\bibliography{references}
\clearpage
\newpage
\appendix
\subsection{Algorithm and Implementation Details}
\label{appendix:algorithm}

\subsubsection{Observations}\hfill

The robot agent first perceives a point cloud $\mathcal{X} \in \mathcal{R}^{M\times3}$ for the entire scene by stacking multiple cameras' views together, where $M$ is the number of points in the raw point cloud observations.
Our method assumes that we pre-process the scene by segmenting the object point cloud $\mathcal{X}^{obj}$ from the background point cloud $\mathcal{X}^b$; details of the segmentation process are listed in Appendix~\ref{appendix:simulation_setup} and Appendix~\ref{appendix:real-world setup}. After segmentation, we downsample $\mathcal{X}^{obj}$ and $\mathcal{X}^b$ with voxel sizes of 1cm and 2cm respectively. We then randomly sample $400$ and $1000$ points from $\mathcal{X}^{obj}$ and $\mathcal{X}^b$ respectively. 

To make our policy also goal-conditioned, we append the goal information into the observation as ``goal flow" 
in which we compute the per-point correspondence from the current object point cloud to the goal object point cloud. Specifically, for each point $x_i$
in the object point cloud, the goal flow is $\Delta x_i = x_i^g - x_i$, where $x_i^g$ is a corresponding point of $x_i$ in the goal point cloud. In the simulation, we use the ground-truth point correspondence given the object pose and the goal pose. In the real world, we use point cloud registration to align the observation to the goal (see Appendix~\ref{appendix:real-world:observation}).

Therefore, the entire observation space $(o_p, o_g, o_m)$ of our robot agent includes three parts: the point cloud $o_p$ representing the 3D position $(x,y,z)$ of the points (3-dimensions per point), the goal flow $o_g$ indicating the flow from the current object point cloud to the point cloud of the object in the goal pose (3-dimensions per point), and the segmentation mask $o_m$ (1-dimension per point).

\vspace{1em}
\subsubsection{Primitive Implementation Details}\hfill\\
\indent We have five generic motion primitives that can be used strategically and collectively to solve long-horizon manipulation tasks. The details of the actual execution of those primitives are listed below.

\textbf{Poke:}
The Poke primitive is parameterized by a location parameter $a^{loc} \in \mathcal{X}^{obj}$ and a motion parameter $a^m=(x^m, y^m, z^m, \theta_x, \theta_y) \in (-1,1)^5$, described below. The gripper first estimates the surface normal $a^{norm}$ of the object at $a^{loc}$ and then goes to the pre-contact location $a^{loc}_{pre} = a^{loc} + a^{norm} \times d_1$, for a hyperparameter $d_1= 0.04$ . After reaching the pre-contact location, the gripper moves to the actual contact location $a^{loc}$ with a rotation $\theta = arctan(\frac{\theta_x}{\theta_y})$ along the z axis. In the last step, the robot moves a delta position $( x^m, y^m, z^m)$ from the contact location. After the poking motion, the gripper returns to the reset pose before it captures the next step observation. 

\textbf{Grasp:}
The Grasp primitive is parameterized by a location parameter $a^{loc} \in \mathcal{X}^{obj}$ and a motion parameter $a^m = (\theta_x, \theta_y) \in (-1,1)^2$. The robot opens the gripper and goes to a pre-contact location $a^{loc}_{pre} = x^{loc} + (0, 0, d_2)$ above the actual location ($d_2=0.1 m$) with a gripper orientation $\theta = arctan(\frac{\theta_x}{\theta_y})$ around the z axis. Then the gripper goes down to the actual contact location and closes the gripper. After grasping, the gripper moves upward in the z axis by $d_3 = 0.15 m$. 

\textbf{Move to:}
The Move to primitive is parameterized by a location parameter $a^{loc}\in \mathcal{X}^b$ and a motion parameter $a^m= (x^m, y^m, z^m, \theta_x, \theta_y) \in (-1,1)^5$. The policy chooses a background point $a^{loc}$ at which to place the object.  Since the gripper is grasping the object, there is an offset between the gripper position and the placed point. Therefore, the actual location the gripper moves to is $a^{loc} + d_4 \cdot (x^m, y^m, z^m)$, where $(x^m, y^m, z^m)$ is a learned offset and $d_4$ is the maximum dimension of the object. During the movement, the gripper also rotates to the orientation $\theta = arctan(\frac{\theta_x}{\theta_y})$ in the z axis.

\textbf{Move delta:}
The Move delta primitive is parameterized by a location parameter $a^{loc} \in \mathcal{X}^b$ and a motion parameter $a^m= (x^m, y^m, z^m, \theta_x, \theta_y) \in (-1,1)^5$. The gripper moves a delta position $(x^m, y^m, z^m)$ with a gripper rotation $\theta = arctan(\frac{\theta_x}{\theta_y})$ about the z axis.

\textbf{Open gripper:}
The Open gripper primitive has no parameters and the selected location $a^{loc}$ also doesn't influence the action. It is an atomic robot action which opens the gripper to the full extent. 

\vspace{1em}
\subsubsection{Baseline Implementation} \hfill

This section provides details of the baselines' key implementation features.  Table~\ref{tab:action_representation_feature_comparisons} summarizes the main differences between our  method and the baseline approaches.

\textbf{P-DQN~\citep{xiong2018parametrized}.} P-DQN uses parameterized primitives, similar to our method, but it lacks spatial grounding in its primitive location selection. In our implementation of this baseline, we processes the point cloud input to derive a global critic feature $f_k$ and a global actor feature $f^a_k$ for each primitive $k$ using a classification-style  network. P-DQN predicts $K$ vectors of primitive parameters and $K$ scores, corresponding to the $K$ primitives. In contrast to our method, for each vector of primitive parameters, P-DQN predicts additionally three dimensions as the regressed location, which are mapped to the predicted Area of Interests (as explained in Section~\ref{sec:regressed location mapping}). P-DQN then selects the primitive with the highest score during inference or samples from the softmaxed scores during exploration.

\textbf{RAPS~\citep{dalal2021accelerating}.} Instead of handling different primitives with separate networks, RAPS extracts a single global actor feature $f^a$ and a single global critic feature $f$ from the input point cloud using a classification-style network. It predicts an action which includes the primitive parameters for all $K$ primitives 
as well as the log-probabilities of executing each primitive.
Similar to P-DQN~\citep{xiong2018parametrized}, RAPS regresses to three dimensions for the primitive location for each primitive, which are mapped to the Area of Interest (as explained in Sec~\ref{sec:regressed location mapping}). RAPS selects the primitive with the highest log-probability in execution and samples from the predicted log-probability during exploration. Note that while our comparison uses TD3~\cite{fujimoto2018addressing} for the baseline to maintain similarity with our method, the original RAPS study experimented with various RL algorithms~\citep{dalal2021accelerating}

\vspace{1em}
\subsubsection{Hyper-parameters}
\hfill

Table~\ref{tab:hyperparameters} lists the training hyper-parameters used in training. 

\begin{table}[h!]
    \centering
    \begin{tabular}{@{}lccc@{}}
\toprule
\textbf{Hyperparameter} & \textbf{Ours} & \textbf{P-DQN} & \textbf{RAPS} \\ \midrule
Target Update Interval & 4 & 1 & 1 \\
Actor Update Interval & 4 & 1 & 1 \\
Learning Rate & 1e-4 & 1e-4 & 1e-4 \\
Batch Size & 64 & 64 & 64 \\
Epsilon Greedy & 0.1 & - & - \\
Action Noise & 0 & 0.1 & 0.1 \\
Exploration Temperature & 0.1 & 0.1 & - \\
Tau & 0.005 & 0.005 & 0.005 \\ \bottomrule
\end{tabular}
    \caption{\textbf{Training Hyper-parameters.}
    }
    \label{tab:hyperparameters}
\end{table}

\vspace{1em}
\subsubsection{Regressed Location Mapping}
\label{sec:regressed location mapping}
\hfill

By spatially-grounding the primitive locations on the input point cloud, our method naturally has an object-centric action space. This allows the more frequent interactions with the objects during exploration. To make it a fair comparison between our method and the baselines, we map the regressed primitive locations predicted by the baseline methods to the Area of Interest (AoI) of each primitive. Specifically, for a raw primitive location prediction $a^loc=(x^m, y^m, z^m) \in (-1,1)^3$, we scale and translate it to the

\begin{enumerate}
    \item the bounding box of the target object for \textbf{object-centric} primitives such as \emph{Poke}, \emph{Grasp}.
    \item the entire workspace for \textbf{background-centric} primitives such as \emph{Move to}.
\end{enumerate}
This change significantly improves the baseline performance, although our experiments show that these baselines still perform significantly worse than our method.

\begin{figure}[t]
    \centering
    \vspace{-9pt}
    \includegraphics[width=\linewidth]{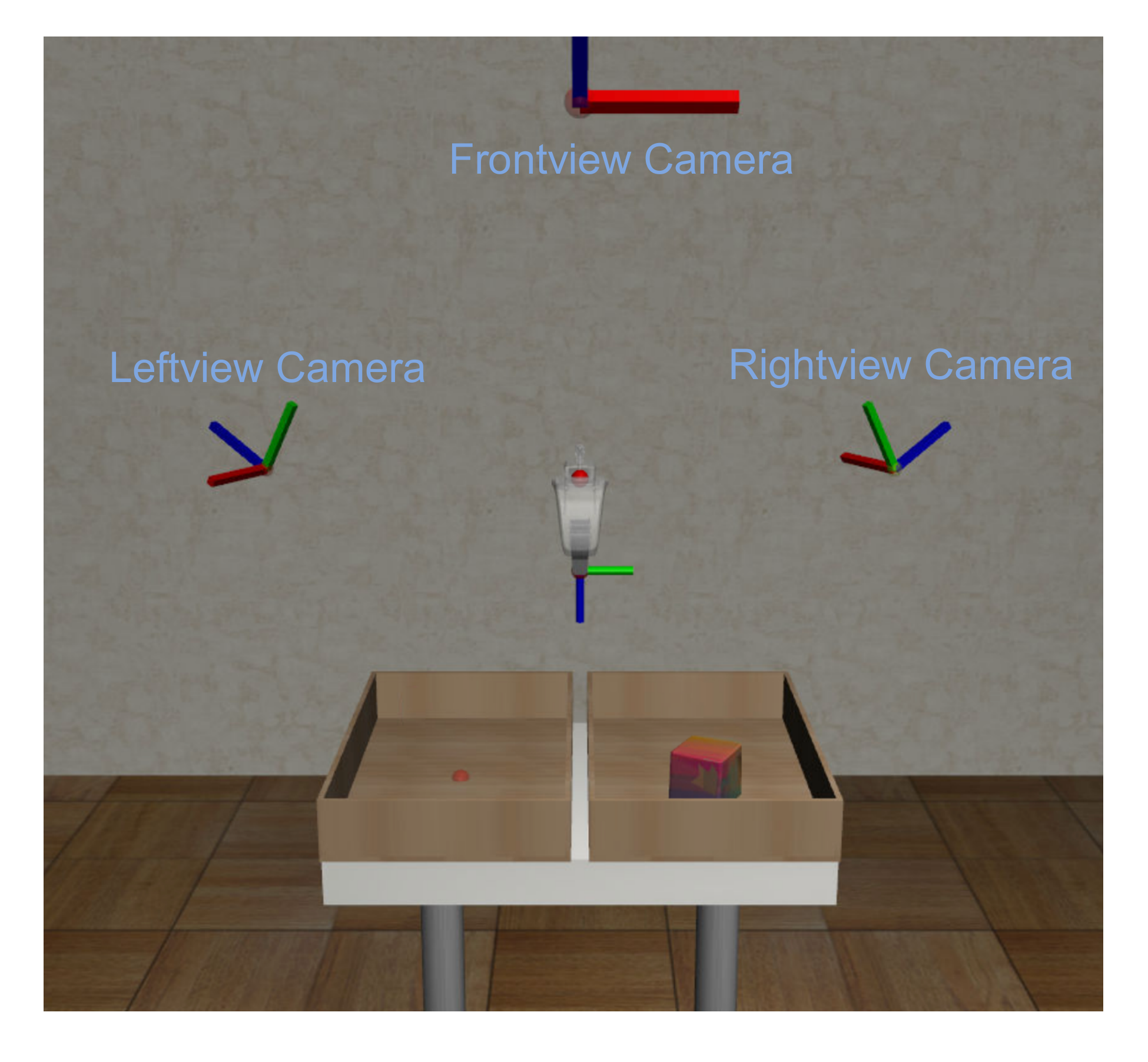}
    \vspace{-22pt}
    \caption{\textbf{Simulation DoubleBin Setup.} The visualization of the simulation environment and the positions of the cameras.}
    \label{fig:simulation_setup}
\end{figure}

\begin{figure}[t]
    \centering
    \includegraphics[width=0.975\linewidth]{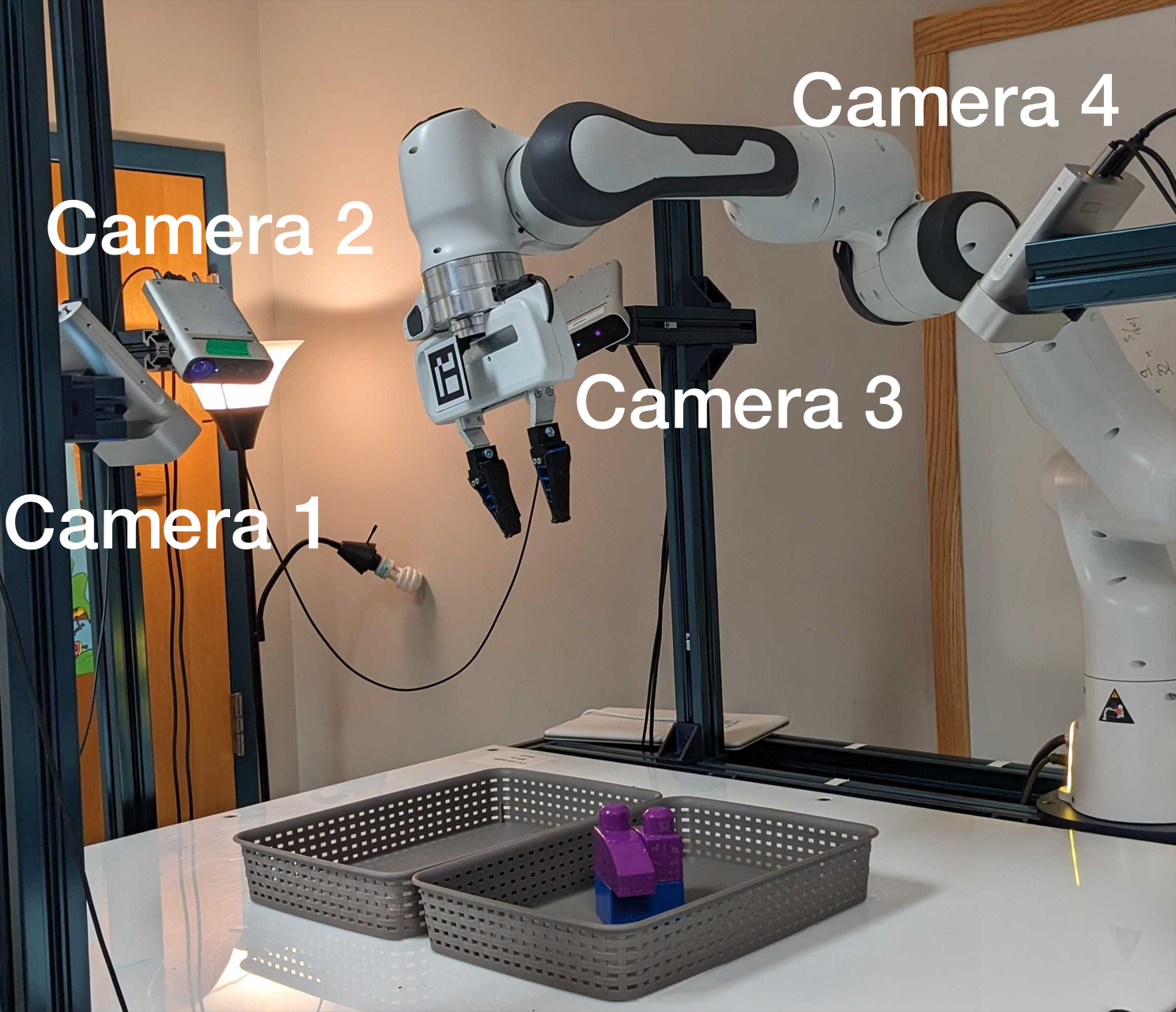}
    \caption{\textbf{Real-world DoubleBin Setup.} The visualization of the real-world DoubleBin environment and the positions of the cameras.}
    \label{fig:real_setup}
\end{figure}

\begin{figure*}
    \centering
    \includegraphics[width=\linewidth]{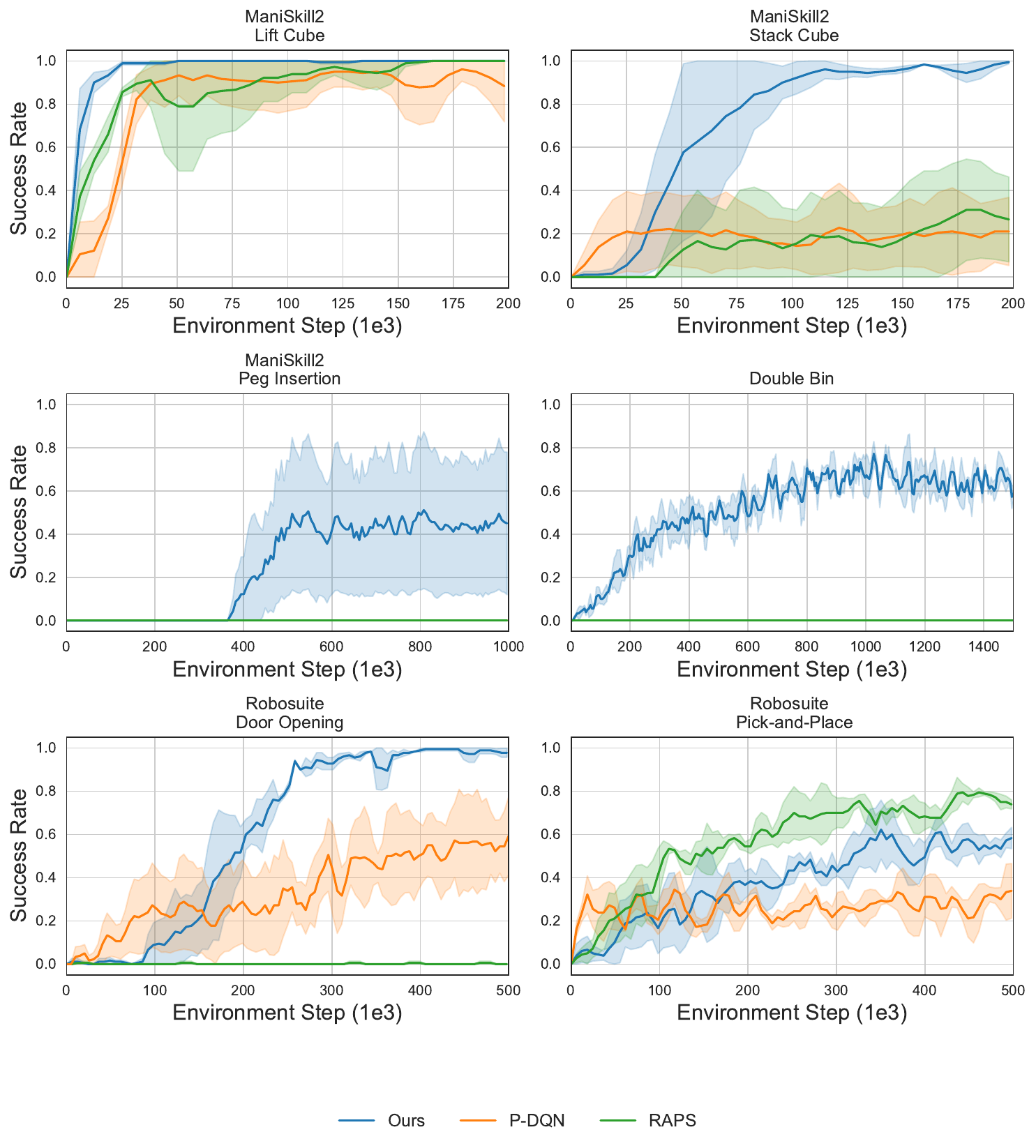}
    \caption{The success rate of six different tasks over environment steps. Each method is averaged over three different seeds and the standard deviation is represented in the shaded area. Our method (\textcolor{blue}{blue}) consistently outperforms both of the baselines in almost all tasks.}
    \label{fig:training_curve}
\end{figure*}

\begin{figure}[ht]
    \centering
    \vspace{2ex}
    \includegraphics[width=\linewidth]{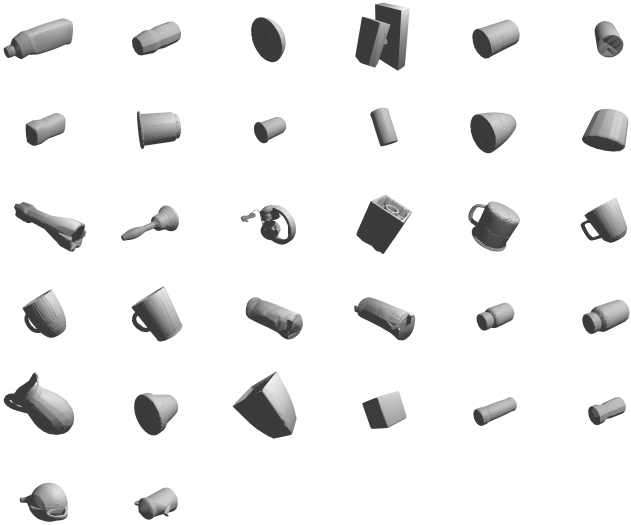}
    \caption{ 32 \textbf{Train} objects used in the training}
    \label{fig:object_train}
\end{figure}
\begin{figure}[ht]
    \centering
     \subfloat[\centering 5 \textbf{Unseen Category} objects]{{\includegraphics[width=.42\linewidth]{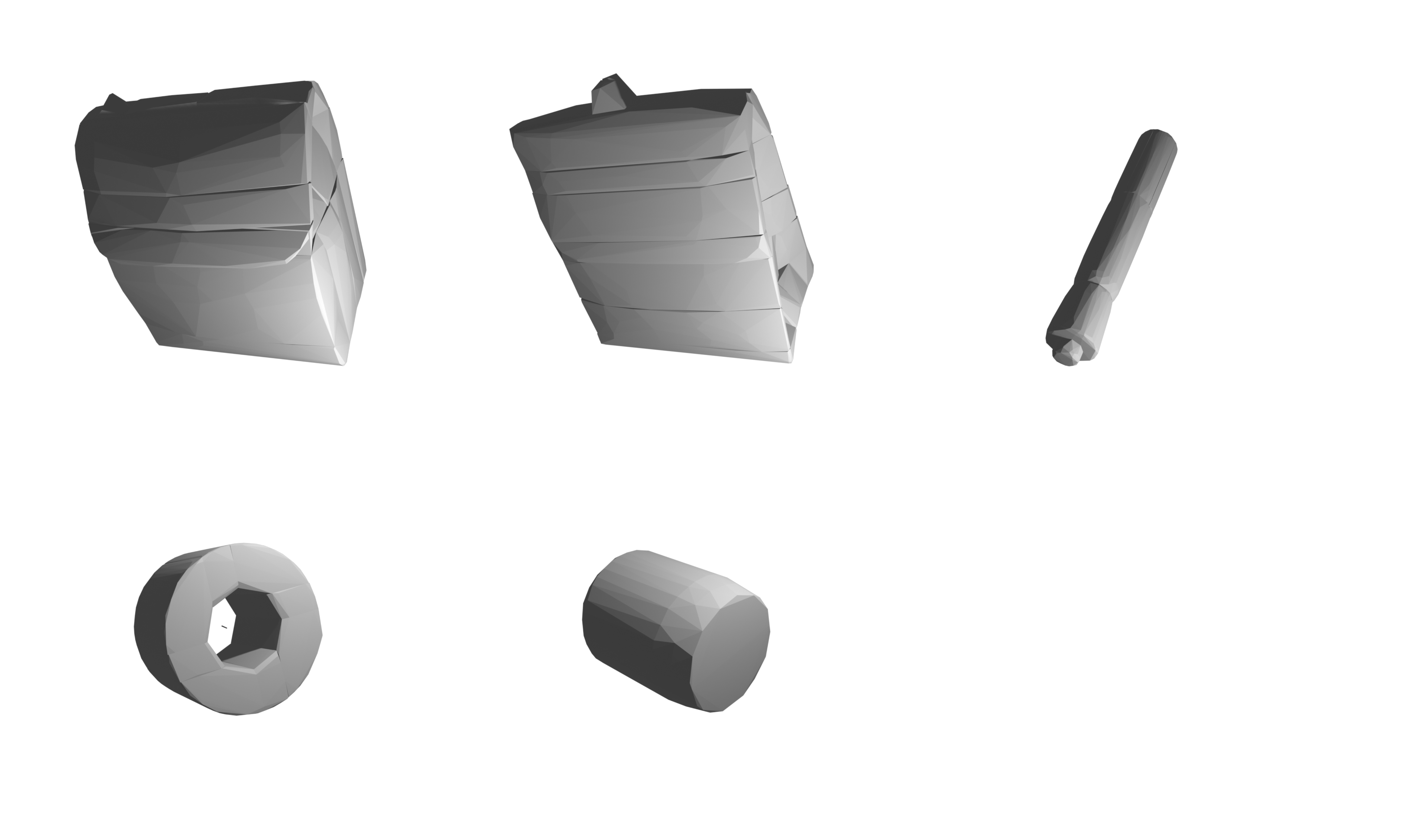} }}%
    \qquad
    \subfloat[\centering 7 \textbf{Unseen Instances (Common Category)} objects]{{\includegraphics[width=.42\linewidth]{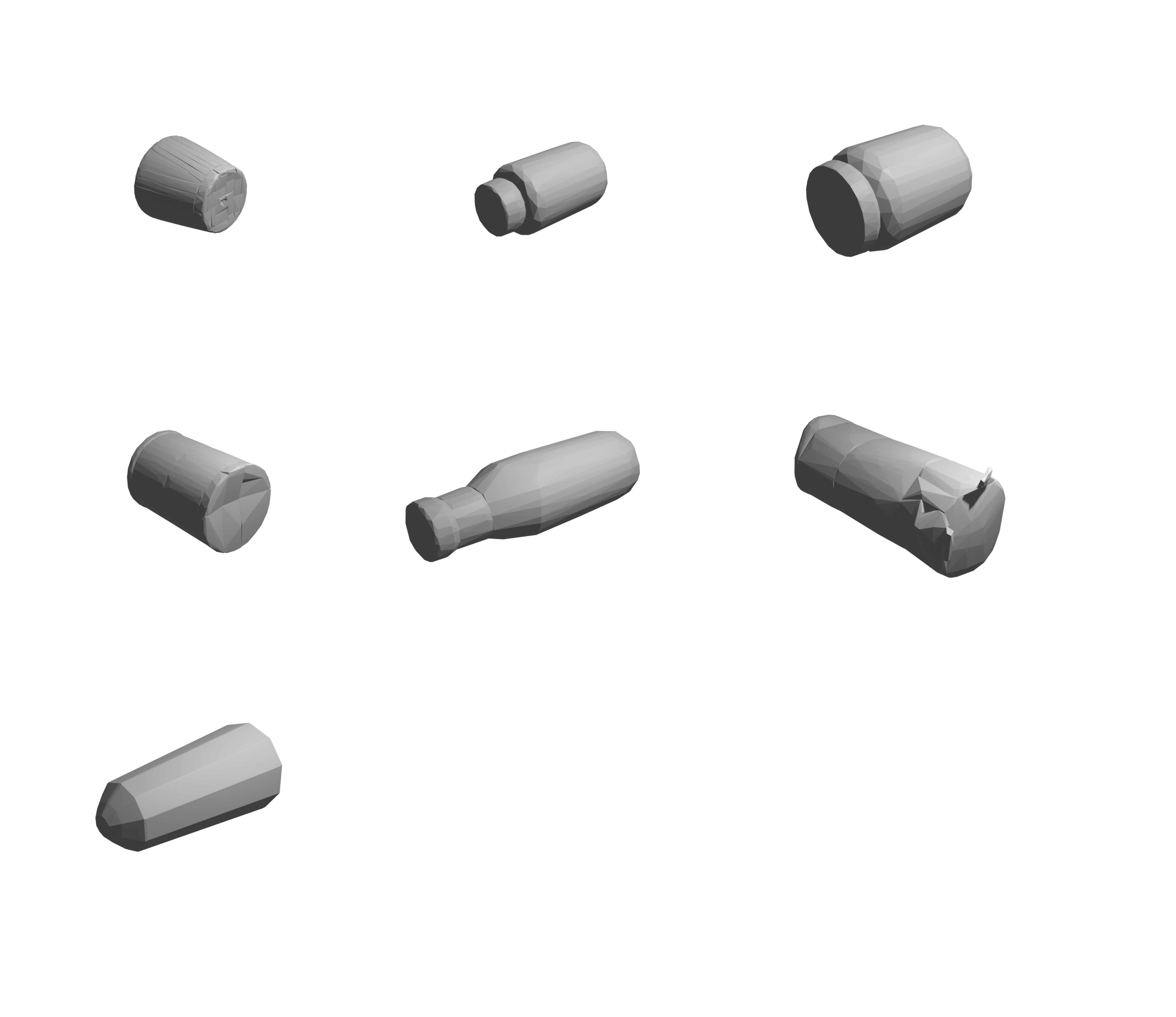} }}%
    \caption{12 Unseen objects used in the evaluation}%
    \label{fig:object_eval}
\end{figure}

\begin{figure}[ht]
    \centering
    \includegraphics[width=\linewidth]{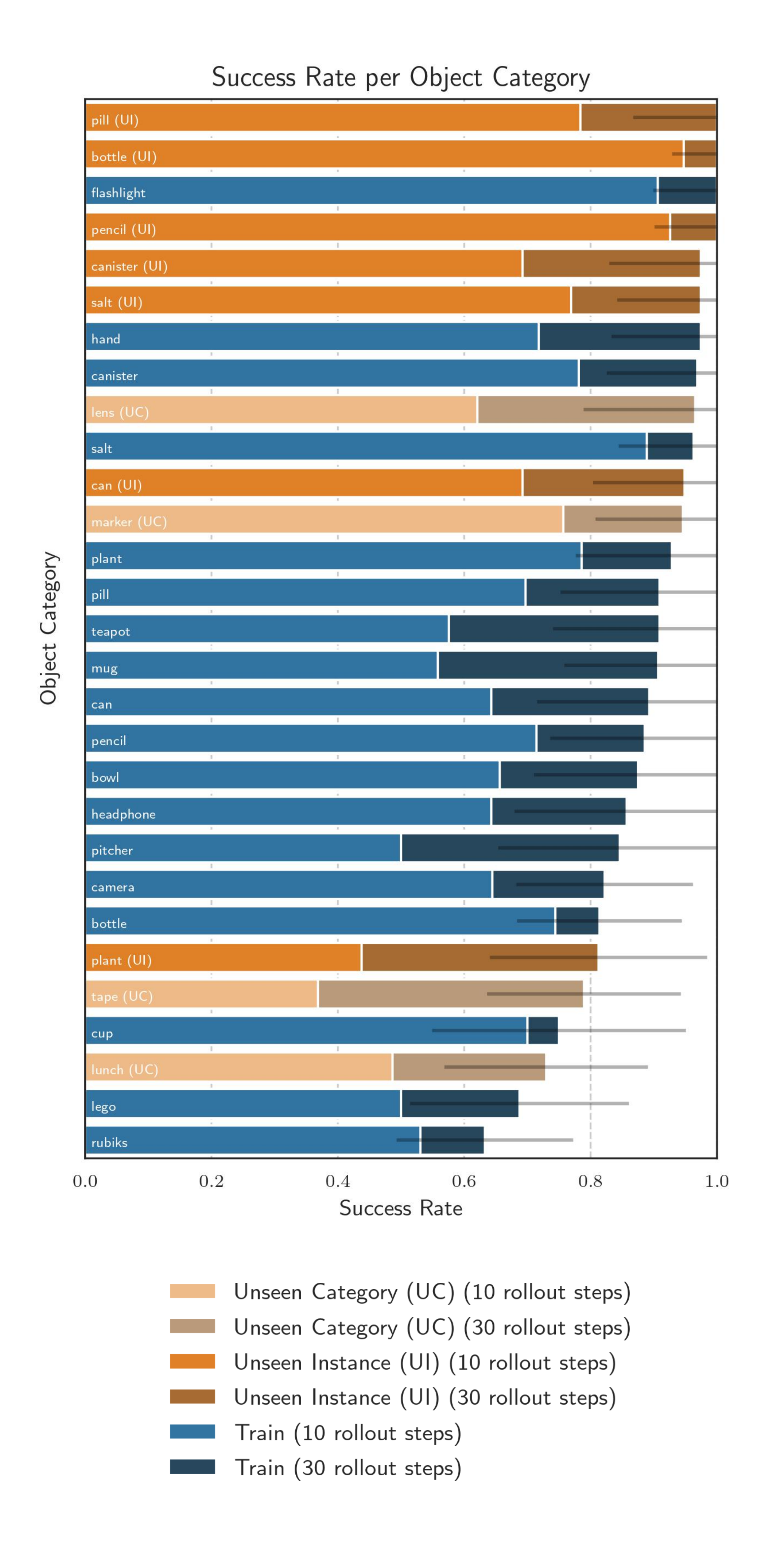}
    \vspace{-40pt}
    \caption{Results breakdown for different object categories with the different maximum episode lengths annotated in the legend as (10) or (30). The Unseen Object Instance (Common Category) (\textcolor{orange}{orange}) has comparable performance with the same object category in the Train (\textcolor{blue}{blue}). The overall success rate of 30 maximum episode steps is higher than the one with 10 maximum episode steps.}
    \label{fig:eval_episode_10}
    \vspace{-0.5cm}
\end{figure}

\subsection{Simulation Experiment Details}
\label{appendix:simulation environments}
Below we describe additional details of our simulation experiments.
\subsubsection{Simulation Tasks}
\label{appendix:simulation_setup}
\begin{enumerate}[label=(\alph*),leftmargin=.5in]
    \item \textbf{\emph{Lift Cube (ManiSkill)}: } We use the Lift Cube task from ManiSkil2~\cite{mu2021maniskill}. The goal of this task is to lift the cube to a goal height of $0.2m$. The initial cube position is uniformly sampled from $[-1, 1]^2$ with a rotation uniformly sampled from $[0, 2\pi]$. The reward function is composed of a reaching reward, grasping reward and lifting reward (see the Maniskill2 documentation for details). We substract ManiSkill2's original reward function by its max value such that the returned reward is always negative, thus encouraging the agent to achieve the success condition as soon as possible  thereby ending the episode. 
    \item \textbf{\emph{Stack Cube (ManiSkill)}: }
    We use the Stack Cube task from ManiSkil2~\cite{mu2021maniskill}. The goal of the task is to pick up a red cube and place it onto a green one. When the red cube is placed on the green cube and not grasped by the robot, the episode is a success (see the ManiSkill2 documentation for details). Similarly, we substract ManiSkill2's original reward function by its max value such that the returned reward is always negative. 
    \item \textbf{\emph{Peg Insertion (ManiSkill)}: } This task is a modified version of the  Peg Insertion Task in ManiSkill2. The goal is to insert a peg into the horizontal hole in a box. We slightly simplify the original task by removing any randomization of the hole's location and peg's initial pose. We also marginally enlarge the hole by adding $1cm$ clearance. Similarly, we substract ManiSkill2's original reward function by its max value such that the returned reward is always negative.
    \item 
    \textbf{\emph{Door Opening (Robosuite)}}:  The task is borrowed from Robosuite Door Opening task. We are using the original dense stage reward from Robosuite. Since there is no specific goal pose for this task, the goal pose is the same as the object pose. 
    \item 
    \textbf{\emph{Pick-and-Place (Robosuite)}}: The task is borrowed from Robosuite Pick-and-Place task. We are using the original dense stage reward from Robosuite. 
    The object is randomly chosen at the beginning of each episode. The original task does not specify a desired goal pose so we choose the center of the container as the desired position and the same orientation of the object as the initial state.
    
    \item \textbf{\emph{DoubleBin Task}: }
We build the customized DoubleBin task environment in Robosuite~\cite{robosuite2020}. The environment has two bins on a table and three cameras, looking over the bins from the left, the right and the front, as shown in the Figure~\ref{fig:simulation_setup}. The size of the bin is $40cm \times 24cm \times 6cm$ and the distance between the centers of the bins is $13.5cm$. For each camera, we record the depth image and project all the points back to the tabletop frame, which has an origin at the middle between the two bins. In Robosuite, we have access to ground-truth segmentation labels of the object and we use these labels to compute a segmentation mask for all the points. We then combine the points from all of the cameras to obtain the point cloud observation. Each episode, an object is randomly chosen from our dataset and loaded into the environment. The object is dropped from the air above one of the bins, after which it reaches a stable initial pose. The reward function for training our RL policy is $r_t = - \frac{1}{N}\sum_{i=1}^{N} ||x_i^g - x_i||$ where $||\cdot||$ is the L2 norm, $N$ is the total number of points on the object point cloud, $x_i$ is a point in the point cloud of the current object pose, and $x_i^g$ is a corresponding point in the point cloud of the goal pose.

The goal poses in the \emph{DoubleBin Task} are sampled by dropping the objects from a certain height and waiting until they stabilize.

The task is considered a success only when $r_t > -0.03$, which is equivalent to $\frac{1}{N}\sum_{i=1}^{N} ||x_i^g - x_i|| < 0.03$m.

\end{enumerate}

\vspace{1em}
\subsubsection{Training and Evaluation Details}
\hfill

In this section, we include the training curves of our methods and baselines over six different task variants. Figure~\ref{fig:training_curve} shows the success rate of the task with regard to the environment interaction steps. Here every environment interaction step refers to one primitive step in the environment which may involve several low-level atomic steps to complete. For every method, we run the experiment across three training seeds. The variance of the seeds is plotted in the figure with the shading area.

We first fill the replay buffer with trajectories by performing 1e4 random actions. Then we start training our policy. During training, we evaluate the current policy every 5e3 environment interaction steps over 20 episodes and report the average success rate across those episodes. 

\vspace{1em}

\subsubsection{Object Dataset Processing and Visualization}\hfill\\
\indent The object dataset we are using is from \citet{liu2022structdiffusion}. We use convex decomposition for the objects and generate watertight mesh following  \citet{zhou2023hacman}. We also scale the objects so that the maximum object dimension is $10$cm. During training, we randomly sample an object and apply additional proportional scaling to all the dimensions within a range of $[0.8, 1.2]$ to simulate objects of different sizes. We also follow the procedure from  \citet{zhou2023hacman} to filter out the objects that have some simulation artifacts or flat and thin objects that are too difficult to poke. After the filtering, we have 44 objects remaining, including cube, bottle, cup, mug, etc. Those objects are divided into three subsets including 32 objects in \textbf{Train}, 7 objects in \textbf{Unseen Instance (Common Categories)} and 5 objects in \textbf{Unseen Category}. Figure~\ref{fig:object_train} shows the \textbf{Train} objects and Figure~\ref{fig:object_eval} shows the \textbf{unseen} objects we test during evaluation.


\begin{figure*}[t]
    \centering
    \includegraphics[width=\linewidth]{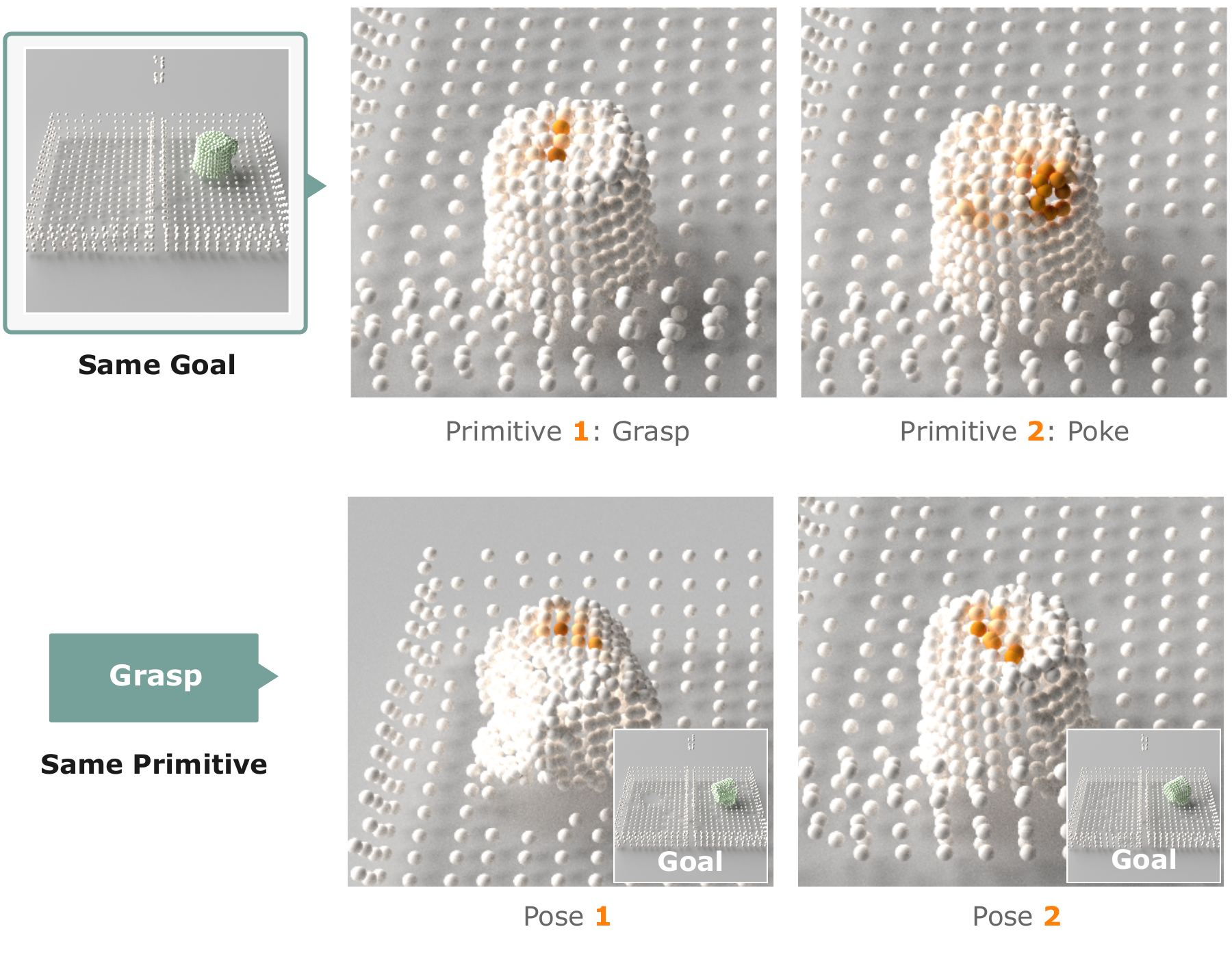}
    \caption{\textbf{Primitive Heatmap}. \textbf{The first row} shows two critic heatmap for two different primitives at the same time step in an rollout. The agent learns to apply different primitives at different regions of the mug, based on its geometric features: \emph{Grasp} needs to be applied to the center of the mug; \emph{Poke} needs to be applied to the side of the mug  to flip it into a more easily graspable pose. \textbf{The second row} shows two critic heatmaps for the same object and the same primitive at two different poses. The agent learns to adapt its grasp location when there is a pose difference of the mug.}
    \label{fig:primitive_heatmap}
\end{figure*}

\vspace{1em}
\subsubsection{Primitive State Estimation}\hfill\\
\indent In simulation, we need to determine whether an object is grasped because certain primitives can only be used when an object is grasped (Move to and Move delta). 
In order to accurately estimate the $\emph{grasped}$ state, we have two conditions. First, we use the Mujoco contact detection to detect if any inner side of the fingertip is in contact with the object. If both inner sides of the fingertip are in contact, we return $\emph{true}$ for the $\emph{grasped}$ state. However, only checking the contact brings some false negative cases because the simulation sometimes cannot detect the contact for some grasping poses due to simulation artifacts. For example, when the gripper grasps the object in corner and it is able to lift the object, the simulation only detects one inner side of the fingertip is in contact of the object. Therefore, to reduce the false negatives, we add the second condition, that is to check if the object is above the table. To be more specific, we check if the object z position is above the table by at least two times of the object's maximum dimension. The final $\emph{grasped}$ state is evaluated to be $\emph{true}$ if either of these two conditions is satisfied. Otherwise, the $\emph{grasped}$ state is $\emph{false}$. 

\vspace{1em}
\subsubsection{Simulation: Additional Evaluation}\hfill\\
\indent In order to have a comprehensive analysis of our method's performance across different geometries and shapes. In this section, we report the breakdown results, i.e., average success rate for each object category. Figure~\ref{fig:eval_episode_10} shows that our method performs consistently well across a large category of objects. However, there are some categories with which our method struggles because of the irregular geometries. 

\begin{figure*}
    \centering
    \includegraphics[width=0.9\textwidth]{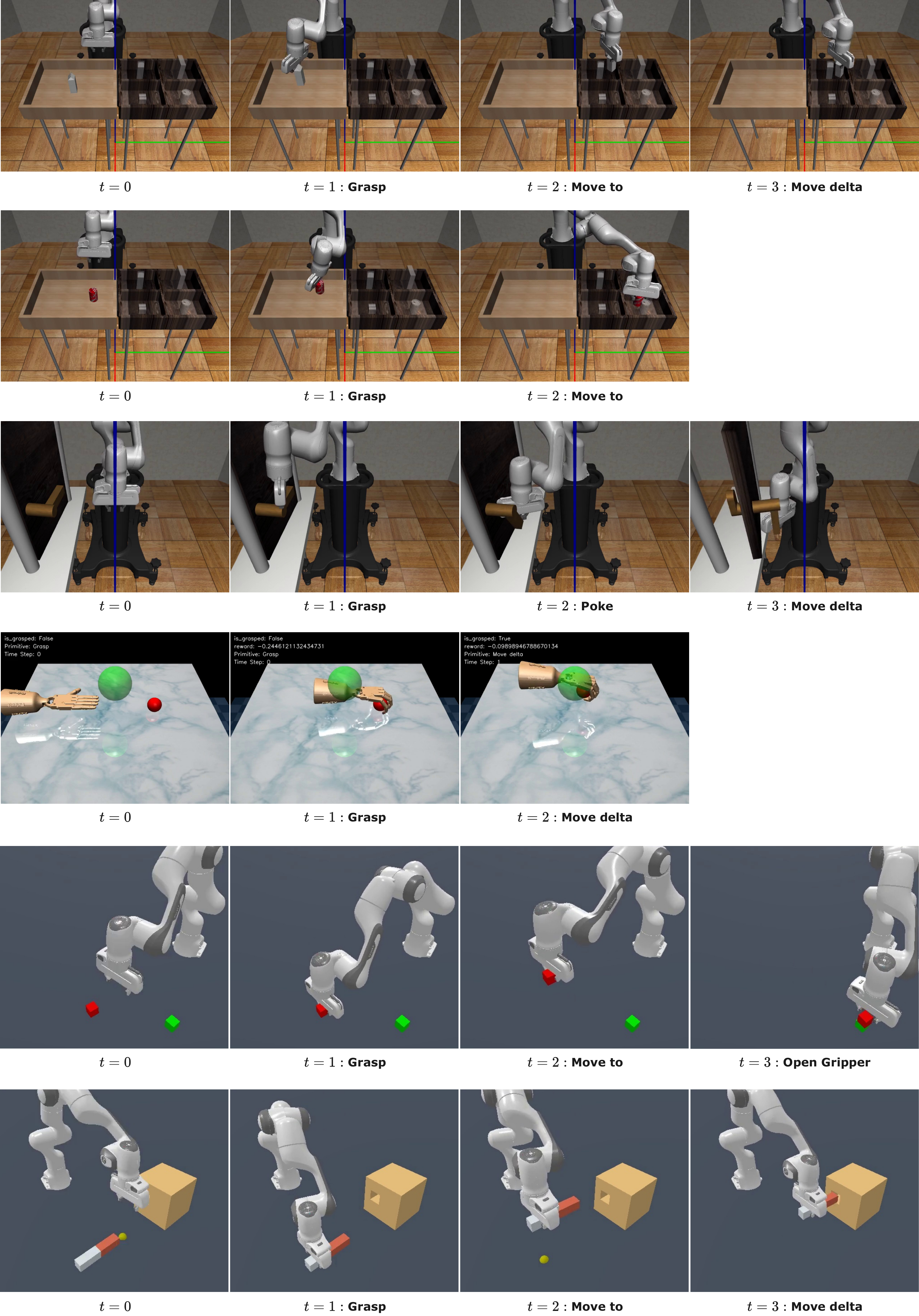}
    \captionsetup{width= 0.9\textwidth}
    \caption{\rebuttal{\textbf{Demonstration of our method's rollouts on Robosuite and Adroit tasks.} The first two rows demonstrate our method performing the Pick-and-Place (RoboSuite) task with different objects. The third row shows our method performing the Door Opening (RoboSuite) task. The forth shows our method performing the Relocate (Adroit) task. The fifth row shows our method performing the Stack Cube (Maniskill2) task. And the last row shows the Peg Insertion (Maniskill2) task.}}
    \label{fig:more-task-rollout}
\end{figure*}

\begin{figure*}[ht!]
    \centering
    \includegraphics[width=\linewidth]{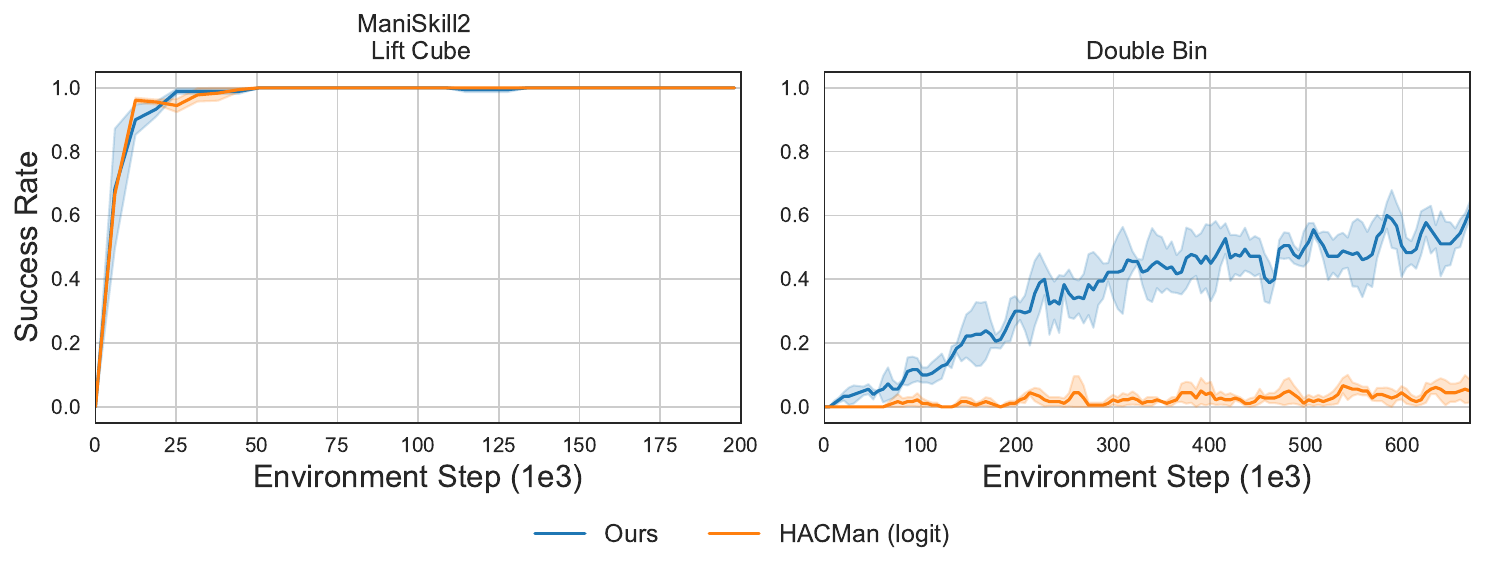}
    \caption{\rebuttal{The success rate of two different tasks over environment steps. Each method is averaged over three different random seeds and the standard deviation is represented in the shaded area. The baseline HacMan(logit) (\textcolor{orange}{orange}) achieves comparable performance on easy task \emph{ManiSkill Lift Cube} but fails to match the performance of our method (\textcolor{blue}{blue}) in more challenging \emph{Double Bin} task.}}
    \label{fig:rebuttal_curve}
\end{figure*}

\vspace{1em}
\subsubsection{\rebuttal{Simulation: Dexterous Hand Task}}
\label{appendix: simulation: additional tasks}
\hfill

 \rebuttalm{To demonstrate that our set of primitives applies to different morphologies of end-effectors, we demonstrate the experimental results on tasks with dexterous hands in simulation apart from the main results we have for grippers.}
 


In the dexterous hand task, we use the \emph{Relocate} task from Adroit simulation benchmark~\cite{rajeswaran2017learning}. The goal is to grasp the red sphere with the ShadowHand and move it to the goal position. The accepted range of the goal is denoted as a large green sphere. We adapt the implementations of the five primitives to be compatible with the robot morphology. When controlling the position of the hand as in \textit{Poke}, \textit{Grasp}, and \textit{Move to}, we align the palm center of the hand to the target position. \textit{Grasp} and \textit{Open Gripper} crunches (all finger joints set to 1) and stretches (all finger joints set to 0) all the fingers respectively. For simplicity, we also remove all the parameters controlling the z-axis rotation. In other words, the hand does not change its orientation during any primitive execution.

Figure~\ref{fig:more-task-rollout} includes an example rollout of our agent. Our method achieves $100\%$ success rate on this dexterous hand task within 20k training steps, proving our method's generality.

\vspace{1em}
\subsubsection{\rebuttal{Simulation: Additional Baseline}}
\label{appendix: simulation: additional baseline}
\hfill

\rebuttalm{
Prior work~\citep{zhou2023hacman} proposes a hybrid discrete-continuous action space for using a single spatially-grounded poking primitive to align the object 6d pose in a single bin. There are multiple ways to extend this single primitive framework to incorporate more primitives to solve a diverse range of general manipulation tasks. In order to demonstrate that designing a framework for multi-primitive setting is non-trivial, we add an additional baseline in this section to show that a naive extension of~\citep{zhou2023hacman} can work on easier tasks like \emph{ManiSkill Lift Cube} task but fail to match our method's performance in more challenging tasks like \emph{Double Bin} task.

We introduce a new baseline, named as HACMan (logit), which extends the continuous action of~\citep{zhou2023hacman} to also include logits. Specifically, the discrete action $a^{loc}$ is to choose a point out of N points in the point cloud and the agent uses that as a location to apply the primitive motion. The continuous action $(a^m_{all}, a^{logit})$ includes the continuous motion parameters $a^m_{all}= (a^m_{grasp}, a^m_{poke}, a^{m}_{move \textunderscore{to}}, a^m_{move \textunderscore{delta}}, a^m_{open \textunderscore{gripper}})$ of all the five primitives and additional 5d logits $a^{logit} \in \mathbb{R}^5$. In execution, the location $a^{loc}$ is chosen based on the Q value of the network. It first chooses the point index $\arg \max_i Q_i(s, a^m_i), 1 \leq i \leq N$ and maps the index to a 3d point location. Then we get the corresponding logits $a^{logit}_i$. The primitive $a^{prim}$ is selected by sampling through the softmax over these logits. The final primitive motion parameter is selected by indexing the motion parameters $a^m_{all}[a^{prim}]$. We conduct the experiments in two different tasks and the results are shown in Figure~\ref{fig:rebuttal_curve}. Compared to our method, which predicts a separate Q value for point and primitive, HACMan (logit) predicts only separate Q value for the point. This structure makes it difficult to learn spatial reasoning with different primitives, leading to its failure on challenging \emph{Double Bin} tasks. }


\color{black}

\vspace{1em}
\subsection{Primitive Heatmap Visualization}
In Figure~\ref{fig:primitive_heatmap}, 
we visualize the spatial Critic maps for different primitives, which vary based on the primitive type and in different regions of an object, based on the geometry of the object. This visualization showcases the agent's capacity for multi-modal reasoning and geometric adaptability in task execution.

\vspace{1em}
\subsection{Real-world Experiments}
\label{appendix:real-world experiments}

\subsubsection{Real World Setup}
\label{appendix:real-world setup}
\hfill

Figure~\ref{fig:real_setup} demonstrates the setup for our real-world DoubleBin experiment. We employ 4 Azure Kinect cameras to capture multi-view point clouds, minimizing the observation occlusion. We use two plastic bins with dimensions similar to the simulation ones, albeit with slight shape differences.

We use a Franka Emika robot equipped with a Franka hand for our experiments. We replace the original Franka hand fingertips with the Festo DHAS-GF-60-U-BU fingertip for improved compliance during contact.

\vspace{1em}
\subsubsection{Observation and Goal Processing}
\label{appendix:real-world:observation}
\hfill

Our input point cloud to the policy contains ``flow", e.g. vectors of correspondences between the observation point cloud and each point's corresponding point in the goal point cloud.  The computation of flow requires us knowing the transformation between the current observation and the goal. In order to estimate this transformation in the real-world, we use point cloud registration. Specifically, this process involves:
\begin{enumerate}[label=(\alph*), leftmargin=.5in]
    \item \textbf{Global registration} using RANSAC with FPFH features.
    \item \textbf{Local refinement} via Point-to-Plane ICP. We only match the object shape, which empirically produces more robust performance than matching both the shape and the color.
\end{enumerate}

We predetermine 4 goal poses for each object by placing them at 2 random positions inside each of the two bins. The robot operates autonomously across episodes without manual intervention. 
When calculating the success rates, we mark episodes with ``fake'' successes (the episode fails but the agent believes it as a success due to point cloud registration failure) as failures.


\vspace{1em}
\subsubsection{Primitive State Estimation}
\hfill

In real-world experiments, we also estimate the primitive state \emph{grasped}. The state is set to $true$ when the camera detects the object's lowest point is at least 4cm above the bin. Conversely, the state switches to $false$ as soon as the gripper releases.

\subsection{\rebuttal{Extended Discussion with Related Work}}
\label{appendix: extended discussion with related work}

\subsubsection{\rebuttal{Compared to~\cite{zhou2023hacman}}}

\rebuttalm{
This previous work only demonstrated a spatially-grounded action space with one type of primitive and one task; we are the first to demonstrate that spatially-grounded manipulation can be extended to a wide range of tasks.\\

To achieve this generality, we design 2 additional spatially-grounded primitives (grasp and move-to) and 2 non-spatially-grounded primitives (open gripper and move-delta) to enable a range of tasks to be achieved. We have presented a set of spatially-grounded primitives that we show to be sufficiently general to be applied to a wide range of tasks and can be adopted by others in the community.

How to spatially-ground each of these primitives is non-obvious and we experimented with different choices of spatial grounding before we found this set of spatially-grounded primitives that work well and cover a range of tasks.

Further, there are multiple ways in which one can imagine extending~\cite{zhou2023hacman} to try to incorporate multiple primitives. In our current approach, we predict a separate Q-value for each point and each primitive, and we choose the point and primitive combination with the highest Q-value. We have added an experiment to compare this approach to an alternative: Similar to RAPS~\cite{dalal2021accelerating}, we extend the actor’s action space to include the log probability (logit) of selecting each primitive; we can treat the logits as a continuous action output which we update using reinforcement learning (e.g. TD3) and then select the action based on a softmax over these logits. This second approach is similar to the approach used in RAPS and this baseline can be viewed as a spatially-grounded variant of RAPS. We conduct experiments to test this approach, as shown in the Appendix~\ref{appendix: simulation: additional baseline}. The results show that while the alternative method solves the easier tasks, our method achieves significantly better performance on the more challenging double-bin task.}

\vspace{1em}
\subsubsection{\rebuttal{Compared to~\cite{feldman2022hybrid}}}

\rebuttal{
Our method has additional primitives that enable our method to achieve a wide range of tasks, compared to the single task shown in~\cite{feldman2022hybrid}.  As mentioned above, designing a set of spatially-grounded primitives that could achieve a wide range of tasks was not straight-forward in our experience. \\
For example, the method in~\cite{feldman2022hybrid} only includes 2 primitives, referred to as “shift” and “grasp” (which are similar to our “poke” and grasp” primitives).  These primitives would not be capable of achieving our double-bin task which requires placing an object into a specific 6D pose.  For this task, we needed the additional primitives that we included: “move to”, “move delta”, and “open gripper”. Other differences compared to~\cite{feldman2022hybrid} include:
}

\begin{enumerate}
    \item \rebuttal{\textbf{2D vs. 3D}. The previous work~\cite{feldman2022hybrid} provides a solution based on the assumption that their spatial grounding can be represented in a top-down 2D pixel space, while we provide a generic solution for manipulation in full-3D-space. The 2D representation limits both the set of available locations and the action flexibility: the set of points that can be selected in~\cite{feldman2022hybrid} are limited to only those visible from a top-down camera, whereas our method can select any visible point on the object surface (such as on the sides of an object); furthermore, the 2D pushing actions used in the previous work are insufficient for more dexterous non-prehensile manipulation motions such as flipping an object, which require 3D pushing actions as we use in our method.}

    \item \rebuttal{\textbf{Limited horizon}. The previous work~\cite{feldman2022hybrid} assumes their task can be completed within 2 primitive steps. It requires a separate value function for each step, which is not scalable for longer horizon-tasks like our Double-Bin task. Additionally, the limited two-step horizon restricts the exploration of multiple solution modalities, as it yields only one possible solution.}

    \item \rebuttal{\textbf{Limited experiments}.}
    \begin{itemize}
        \item \rebuttal{Tasks diversity. The prior work~\cite{feldman2022hybrid} focuses on how to pick up objects, with two specialized primitives designed for this task (grasping and 2D shifting). We focus on how an agent can discover longer-horizon strategies using a diverse set of primitives on more general manipulation tasks. }

        \item \rebuttal{Object generalization: \cite{feldman2022hybrid} shows their method working on limited objects geometry (cubes, spheres, and cylinders). Prior work \cite{zhou2023hacman} has shown that the difficulty is much lower when there is limited geometric diversity. In contrast, our method maintains good performance across diverse and even diverse unseen geometries.}

        \item \rebuttal{The prior work \cite{feldman2022hybrid} does not show any real-world experiments and does not explore how the method can be transferred to real-world.}
    \end{itemize}
\end{enumerate}

\end{document}